\documentclass{article}
\usepackage{titlesec}
\usepackage{cite}
\usepackage{enumitem}
\usepackage{authblk}
\usepackage{geometry}
\usepackage{subfigure}
\usepackage{multirow}
\usepackage{booktabs}
\usepackage{amsmath}
\usepackage{amssymb}
\usepackage{amsfonts}
\usepackage{algorithm}
\usepackage{stmaryrd}
\usepackage{algpseudocode}
\usepackage{array}
\usepackage{float}
\usepackage{textcomp}
\usepackage{subcaption}
\usepackage{ccicons}
\usepackage{arydshln}
\usepackage{hyperref}
\usepackage{graphicx}
\usepackage{doi}

\title{\textbf{Opus} \vspace{0.5em} \\ \Large{\fontseries{bx} A Prompt Intention Framework for Complex Workflow Generation}}

\author{
\vspace{1em}

\begin{minipage}[t]{0.25\textwidth}
    \centering
    \begin{tabular}[t]{c}
        \small{\textbf{Théo Fagnoni}} \\ \vspace{-0.7em}
        \scriptsize{Member of Technical Staff} \\
        \scriptsize{AppliedAI} \\
    \end{tabular}
\end{minipage}%
\begin{minipage}[t]{0.25\textwidth}
    \centering
    \begin{tabular}[t]{c}
        \small{\textbf{Mahsun Altin}} \\ \vspace{-0.7em}
        \scriptsize{Member of Technical Staff} \\
        \scriptsize{AppliedAI} \\
    \end{tabular}
\end{minipage}%
\begin{minipage}[t]{0.25\textwidth}
    \centering
    \begin{tabular}[t]{c}
        \small{\textbf{Chia En Chung}} \\ \vspace{-0.7em}
        \scriptsize{Member of Technical Staff} \\
        \scriptsize{AppliedAI} \\
    \end{tabular}
\end{minipage}%
\begin{minipage}[t]{0.25\textwidth}
    \centering
    \begin{tabular}[t]{c}
        \small{\textbf{Phillip Kingston}$^*$}\\ \vspace{-0.7em}
        \scriptsize{Member of Technical Staff} \\
        \scriptsize{AppliedAI} \\
    \end{tabular}
\end{minipage}%
\vspace{\baselineskip}

\hspace*{0.125\textwidth}%
\begin{minipage}[t]{0.25\textwidth}
    \centering
    \begin{tabular}[t]{c}
        \small{\textbf{Alan Tuning}} \\ \vspace{-0.7em}
        \scriptsize{Research Affiliate} \\
        \scriptsize{AppliedAI} \\
    \end{tabular}
\end{minipage}%
\begin{minipage}[t]{0.25\textwidth}
    \centering
    \begin{tabular}[t]{c}
        \small{\textbf{Dana O. Mohamed}}\\ \vspace{-0.7em}
        \scriptsize{Member of Technical Staff} \\
        \scriptsize{AppliedAI} \\
    \end{tabular}
\end{minipage}%
\begin{minipage}[t]{0.25\textwidth}
    \centering
    \begin{tabular}[t]{c}
        \small{\textbf{Inès Adnani}} \\ \vspace{-0.7em}
        \scriptsize{Research Affiliate} \\
        \scriptsize{AppliedAI} \\
    \end{tabular}
\end{minipage}%
\hspace*{0.125\textwidth}
\vspace{1em}
}

\date{\normalsize{30 June 2025}}

\begin{document}

\maketitle
\begin{center}
    \ccbyncsa \\
    \vspace{0.3em}
    \footnotesize{This work is licensed under a Creative Commons Attribution-Noncommercial-ShareAlike 4.0 International License (CC BY-NC-SA 4.0)}
\end{center}

\footnotetext{\text{$^*$ Corresponding author: phillip.kingston@opus.com}}

\vspace{1em}

\begin{abstract}

\noindent 
This paper introduces the Opus Prompt Intention Framework, designed to improve complex Workflow Generation with instruction-tuned Large Language Models (LLMs). We propose an intermediate Intention Capture layer between user queries and Workflow Generation, implementing the Opus Workflow Intention Framework, which consists of extracting Workflow Signals from user queries, interpreting them into structured Workflow Intention objects, and generating Workflows based on these Intentions. Our results show that this layer enables LLMs to produce logical and meaningful outputs that scale reliably as query complexity increases. On a synthetic benchmark of 1,000 multi-intent query–Workflow(s) pairs, applying the Opus Prompt Intention Framework to Workflow Generation yields consistent improvements in semantic Workflow similarity metrics. In this paper:
\begin{enumerate}
    \item We introduce the Opus Prompt Intention Framework by applying the concepts of Workflow Signal and Workflow Intention to LLM-driven Workflow Generation.
    \item We present a reproducible, customizable LLM-based Intention Capture system to extract Workflow Signals and Workflow Intentions from user queries.
    \item We provide empirical evidence that the proposed system significantly improves Workflow Generation quality compared to direct generation from user queries, particularly in cases of Mixed Intention Elicitation.
\end{enumerate}
\end{abstract}

\newpage

\section{Introduction}
In the modern business environment, well-structured internal processes form the backbone of operational consistency, efficiency, and accountability. Yet, the quality and clarity of process documentation across organizations varies in completeness, accuracy, and granularity, making automation difficult to scale. As businesses face growing demands for agility, cost reduction, and regulatory compliance, many are turning to AI-driven systems to streamline decision-making and operational execution. However, realizing the full potential of automation requires systems that can understand and act on human intent expressed in natural language. This is particularly challenging when user queries are informal, ambiguous, or contain multiple goals. To address this, we adopt the Opus Workflow Intention Framework—a structured representation of Input, Process, and Output—to bridge the gap between human language and machine-executable Workflows. In this paper, we propose the Opus Prompt Intention Framework that leverages LLMs to extract these Intention objects from user queries and evaluate whether incorporating them explicitly into the Workflow Generation process leads to improved quality. We aim to demonstrate that Workflows generated from Intention objects will be more accurate, interpretable, and aligned with user goals than those generated from raw queries alone. Critically, the Opus Prompt Intention Framework enables scalable Workflow Generation with respect to query complexity. As the complexity of user input increases—whether through multi-objective requests, incomplete specifications, or ambiguous phrasing—the proposed framework maintains high accuracy and consistency, while traditional LLM-based approaches degrade in performance. The Opus Prompt Intention Framework not only enables more reliable automation, but also offers a scalable path for organizations to convert informal, fragmented practices into structured, AI-enhanced Workflows that reflect contemporary best practices.

\paragraph{Definitions}

The Opus Prompt Intention Framework is based on the following concepts introduced in \textit{Opus: A Large Work Model for Complex Workflow Generation} by Fagnoni et al. \cite{opuslargeworkmodel} and \textit{Opus: A Workflow Intention Framework for
Complex Workflow Generation} by Kingston et al. \cite{opusintentionframework}:

\vspace{1em}

\noindent\hangindent=2em\hangafter=0 \textbf{\textbf{Input}:} The dataset initiating a \textbf{Process}, conforming to validation rules and format specifications. \textbf{Input} is multimodal, including structured (e.g. databases, forms) and unstructured (e.g. documents, media) data types such as text, documents, images, audio, and video.

\vspace{1em}

\noindent\hangindent=2em\hangafter=0 \textbf{Process:} A structured sequence of operational steps transforming \textbf{Input} into \textbf{Output}. \textbf{Process} combines automated and manual steps defining start/end conditions, decision points, parallel paths, roles, success criteria, monitoring, metrics, compliance requirements and error handling.

\vspace{1em}

\noindent\hangindent=2em\hangafter=0 \textbf{Output:} The result of a \textbf{Process} operating on \textbf{Input}, meeting predefined quality and business criteria. \textbf{Output} can be tangible (e.g. documents) or intangible (e.g. decisions) and include audit trails of their creation. Supported formats include text and documents.

\vspace{1em}

\noindent\hangindent=2em\hangafter=0 \textbf{Workflow:} A software executable \textbf{Process} as a Directed Acyclic Graph (DAG) of \textbf{Tasks}. \textbf{Workflows} coordinate task execution, manage data flow, and enforce business rules, compliance, and process logic (e.g. conditionals, loops, error handling). \textbf{Workflows} support monitoring, logging, audit, state management, concurrency, adaptive modification, and version control.

\vspace{1em}

\noindent\hangindent=2em\hangafter=0 \textbf{Task:} An atomic unit of work within a \textbf{Workflow}, performing a specific function with defined input/output schemas, objectives, timing constraints, and success criteria. \textbf{Tasks} follow a singular responsibility principle, support automation or manual intervention, and maintain contextual awareness of dependencies. \textbf{Tasks} are auditable by humans or AI agents against their definition.

\vspace{1em}

\noindent\hangindent=2em\hangafter=0 \textbf{Workflow Intention} (referred to as \textbf{Intention}): The alignment of \textbf{Input}, \textbf{Process}, and \textbf{Output} components defining a \textbf{Workflow}'s transformation objective. It specifies how \textbf{Input} is processed to achieve desired \textbf{Output}, incorporating data formats, quality standards, business rules, and constraints. It is determined by interpreting \textbf{Workflow Signals} from direct and indirect sources.

\vspace{1em}

\noindent\hangindent=2em\hangafter=0 \textbf{Workflow Signal:} A discrete informational cue from \textbf{Intention Elicitation} that conveys implicit or explicit information on \textbf{Input}, \textbf{Process} or \textbf{Output} relevant to a \textbf{Workflow}.

\vspace{1em}

\noindent\hangindent=2em\hangafter=0 \textbf{Complete Intention:} A state where sufficient information exists across \textbf{Input}, \textbf{Process}, and \textbf{Output} components for accurate \textbf{Workflow} implementation. \textbf{Incomplete Intentions}, on the other hand, lack clear specifications or operational requirements, hindering execution.

\vspace{1em}

\noindent\hangindent=2em\hangafter=0 \textbf{Intention Elicitation:} User-driven communication (e.g. text-based conversations, interface interactions) that contains \textbf{Workflow Signals} to further articulate \textbf{Workflow Intention(s)}. It captures objectives, constraints, and preferences, distinct \textbf{Input}/\textbf{Output} examples.

\vspace{1em}

\noindent\hangindent=2em\hangafter=0 \textbf{Mixed Intention Elicitation:} A state where the \textbf{Intention Elicitation} describes multiple distinct transformation objectives, requiring separation into \textbf{Workflow Intentions}, in contrast to \textbf{Singular Intention Elicitation}, which expresses a single, well-defined \textbf{Workflow Intention}. Separation improves clarity, maintainability, and preserves \textbf{Workflow} interfaces.

\vspace{1em}

We consider the problem of Workflow Generation from an Intention Elicitation, where a user provides a query describing the Workflow(s) they intend to generate. Two key principles govern the system design for this problem.
First, when a user query does not fully specify a Workflow—formally, when it expresses an Incomplete Intention—the system must be able to reliably identify this condition. Attempting to resolve incomplete specifications directly during Workflow Generation often leads to unreliable or suboptimal results. Instead, the system should proactively flag such incompleteness for user clarification, or attempt to retrieve the missing components from external sources (e.g., knowledge graphs) before proceeding with Workflow Generation. Second, when a user query expresses multiple distinct transformation objectives—a Mixed Intention Elicitation—the system must generate a separate Workflow for each distinct transformation. Generating a single Workflow to satisfy multiple objectives leads to unintended contextual overlaps and reduced accuracy. The system should instead trigger multiple, distinct Workflow Generations, each aligned to a singular objective. These two principles highlight the need for a dedicated preprocessing layer in Workflow Generation. Our results show that the Opus Workflow Intention Framework provides a scalable and reliable foundation for structuring this layer.

\section{Background}

\paragraph{Workflow, Workflow Intention} We adopt the Opus Workflow Framework (Fagnoni et al. \cite{opuslargeworkmodel}) where a Workflow is represented as a Directed Acyclic Graph (DAG) consisting of Input Nodes, Task Nodes, Output Nodes, and Edges that define the Task execution flow from Inputs to Outputs. We generate Workflows at a semantic level: each Node is described using structured semantic features following a predefined schema, typically represented as a JSON object with fields such as name, description, and steps for Task Nodes. The schema is flexible and can be adapted as required, demonstrating the extensibility of the framework. Additionally, we adopt the Opus Workflow Intention Framework (Kingston et al. \cite{opusintentionframework}), which maps business artefacts and Intention Elicitation into Workflow Signals—specifically Input ($i$), Process ($p$), and Output ($o$) elements—and into an Intention Set. In this work, we focus exclusively on Intention Elicitation as user queries, in text format. Since we operate directly at a semantic level using LLMs, each Workflow Signal is represented as a list of strings. This aligns with the Opus Workflow Intention Framework, where Workflow Signals are modeled as vectors and classified against generative families of Input, Process, and Output string elements. In our representation, each string corresponds to a distinct semantic component, enforced through prompt design. The generation of these strings is unconstrained beyond this granularity. Although further classification against generative families is possible, we do not consider this step essential to the core argument of this paper. Intention objects, forming an Intention Set for each Intention Elicitation, are represented as triples $(i,p,o)$, where each element is a list of strings.

\paragraph{Large Language Models (LLMs)}
Recent advancements in LLMs have significantly improved text generation, multimodal reasoning, and processing efficiency. Built on Transformer architectures \cite{attention-is-all-you-need}, LLMs leverage attention mechanisms for scalable parallel computation. Beyond pattern recognition and text generation, modern reasoning-oriented models integrate Reinforcement Learning \cite{deepseek-r1 , gpt4-technical-report, gemini-google, llama3-meta}, Mixture of Experts (MoE) \cite{deepseekv3, llama4-meta, qwen3}, and long-context handling \cite{yi-1.5, mistral-7b, gemma3-google, gpt-neox} to improve logical consistency and decision-making. Leading these developments are OpenAI’s GPT-4o \cite{openai-gpt-4o}, o1 \cite{openai-o1}, o3-mini \cite{openai-o3-mini}, and GPT-4.5 \cite{openai-gpt-4.5}; Anthropic’s Claude Opus  \cite{claude-3-opus}, Sonnet 3.5 \cite{claude-3.5-sonnet} and 3.7 \cite{claude-3.7-sonnet}; and DeepSeek’s R1 \cite{deepseek-r1}. Persistent hallucinations, high computational costs, and training data bottlenecks necessitate robust pipelines and frameworks to improve alignment and enhance reliability in real-world applications. Building on these advancements, LLMs provide a strong foundation for extracting Workflow Signals and generating Workflows from user queries, including complex, multi-intent scenarios. Although current models may struggle with processing multiple Intentions simultaneously, their capacity to parse nuanced language, reason over context, and infer latent structure positions them as well-suited for aligning Workflow Signals—even when such Signals are implicit. This capability enables the construction of modular and interpretable Intention objects that guide Workflow Generation. Whether through direct query-to-Workflow Generation or via structured Intentions, LLMs offer the flexibility and language grounding required to bridge natural language with executable systems.

\newpage

\paragraph{Workflow Quality} The core argument of this paper is that Workflow Generation is significantly improved when guided by explicit Workflow Intentions. To substantiate this claim, we evaluate generated Workflows using two categories of metrics:
(1) semantic and structural metrics, which assess Workflow similarity, and (2) LLM-based metrics,  where LLMs are prompted to evaluate Workflow quality under predefined criteria. For semantic and structural metrics, we adopt metrics used in the Opus Workflow Framework \cite{opuslargeworkmodel}. These include BLEU (Bilingual Evaluation Understudy, Papineni et al. \cite{bleu-score}), a precision-based metric widely used in machine translation; ROUGE (Recall-Oriented Understudy for Gisting Evaluation, Lin et al. \cite{rouge-score}) and its variants—ROUGE-N (N-gram overlap), ROUGE-L (Longest common subsequence), and ROUGE-S (Skip-bigram co-occurrence)—commonly applied in summarization evaluation; METEOR (Metric for Evaluation of Translation with Explicit ORdering, Banerjee et al. \cite{meteor-score}), which extends BLEU by incorporating stemming, synonym matching, and a higher recall weighting; and BERTScore (Zhang et al. \cite{bert-score}), which leverages contextual embeddings from pre-trained language models (specifically BERT) to measure textual similarity. 
For LLM-based evaluation, we adopt the LLM-as-a-Judge framework, where LLMs are used to evaluate model outputs. Gu et al. \cite{survey-llm-as-a-judge} provide a comprehensive overview of this approach, addressing challenges related to bias, robustness, and alignment with human judgments. When properly configured, LLM-based evaluation provides a powerful tool for qualitative assessment of complex Workflows. However, the inherent imprecision of LLM judgments—stemming from variability in model outputs and susceptibility to bias—requires mitigation strategies such as prompt calibration and repeated sampling. Employing multiple models as evaluators can help temper these limitations by diversifying judgment perspectives.

\section{System Overview}
\begin{figure}[H]
    \centering
    \includegraphics[width=\textwidth]{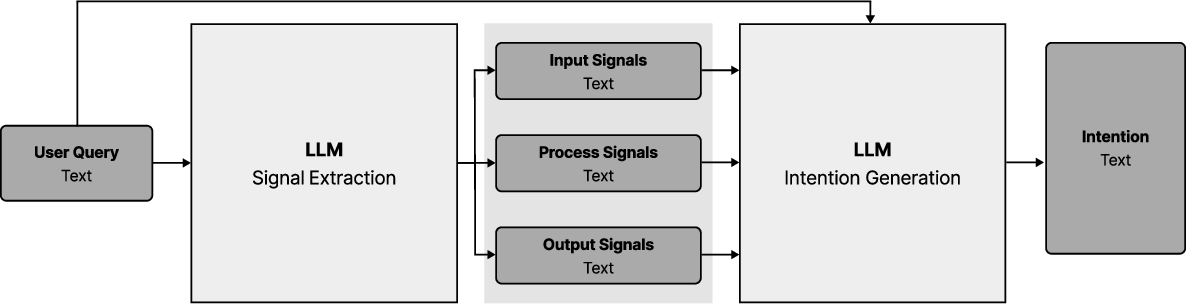}
    \caption{LLM-based Intention Capture}
    \label{fig:llm_solution}
\end{figure}

We implement a reproducible Intention Capture system using LLMs, designed to be model-agnostic and demonstrate that even a basic Intention Capture improves Workflow Generation unequivocally. The system consists of two LLMs, as illustrated in Figure \ref{fig:llm_solution}: the first extracts Input, Process, and Output Signals from an Intention Elicitation; the second generates the Intention Set, where each Intention object comprises aligned Input, Process, and Output Signals. In this context, a Signal is a list of granular string elements within its category (Input, Process, or Output), mirroring the classification of encoded artefacts into predefined elements of families as defined in the Opus Workflow Intention Framework \cite{opusintentionframework}. We evaluate the accuracy of the generated Signals and Intention Set using loss functions analogous to those employed in the Opus Workflow Intention Framework.

\section{Data and Methodology}
Starting with 1,000 registered services spanning 100 industries, we construct a Process Pool $\text{P}_g$ consisting of semantic Process string elements. For each Process element, we generate corresponding Input and Output elements, forming a Complete Intention object $\gamma = (i, p, o)$, where each component is a list containing at least one item. This construction yields the Input Pool $\text{I}_g$ and the Output Pool $\text{O}_g$. We then define the Intention Pool $\Gamma_p$, where each Intention is composed of three lists, each populated with elements from $\text{I}_g$, $\text{P}_g$ and $\text{O}_g$. \\

For each Intention $\gamma = (i, p, o)$, we generate a corresponding Intention Elicitation (as a user query) $\mathcal{U}$—a string crafted from $i, p, \text{ and }o$, ensuring these components are explicitly stated and incorporated. To increase linguistic entropy and make the mapping non-injective, we introduce controlled variations in phrasing that preserve semantic content. This simulates user query diversity—akin to adding noise to a signal without losing information. From $\mathcal{U}$, we generate a Workflow $\text{W}$, represented as a string (structure detailed in Appendix \ref{app:prompts}, Figure \ref{fig:workflow-generation-with-intention}). To simulate multi-intent scenarios, we sample a variable number $n_k \in \llbracket 1, 10 \rrbracket$ of Intention Objects from $\Gamma_p$ to form an Intention Set, comprising the Reference Intention Sets:

\begin{equation}
    \Gamma^{*}  = \{\Gamma^{*, k}\}_k  \quad\text{where} \quad \Gamma^{*,k} = \{(i^{k}_j,p^{k}_j,o^{k}_j)\}^{n_k}_{j=1}
\end{equation}

Each $\Gamma^{*,k}$ is used to construct:

\begin{enumerate}
\item A set of Singular Intention Elicitation (user query) $\big\{\mathcal{U}^k_j\big\}_{j=1}^{n_k}$, where each query corresponds to an Intention object in $\Gamma^{*,k}$.
\item A combined Mixed Intention Elicitation $\mathcal{U}_m^k$, synthesized by aggregating all individual user queries corresponding to the Intention objects within $\Gamma^{*,k}$.
\item A Reference Workflow Set $\mathcal{W}^{*,k}=\big\{\text{W}^k_j\big\}_{j=1}^{n_k}$, where each Workflow is generated from a Singular Intention Elicitation.
\end{enumerate}

The Sample Set is defined as:
\begin{equation}
\mathcal{S} = \Bigl\{
    \Bigl(
        \Gamma^{*, k}, \;
        \bigl\{\mathcal{U}^k_j\bigr\}_{j=1}^{n_k}, \;
        \bigl\{\text{W}^k_j\bigr\}_{j=1}^{n_k}, \;
        \mathcal{U}_m^k
    \Bigr)
\Bigr\}_k, \quad \forall k, \; n_k \in \llbracket 1, 10 \rrbracket
\end{equation}

Each sample $k$ consists of:
\begin{enumerate}
    \item A Mixed Intention Level $n_k\in\llbracket1, 10\rrbracket$, defined as the number of Intention objects used to generate the Mixed Intention Elicitation.
    \item A Reference Intention Set $\Gamma^{*, k} = \big\{(i^k_j, p^k_j, o^k_j)\big\}_{j=1}^{n_k}$ composed of Complete Intention objects.
    \item Singular Intention Elicitations (user queries) $\big\{\mathcal{U}^k_j\big\}_{j=1}^{n_k}$, each corresponding to an Intention in $\Gamma^{*, k}$.
    \item A Reference Workflow Set $\mathcal{W}^{*,k}=\big\{\text{W}^k_j\big\}_{j=1}^{n_k}$, where each Workflow is generated from its corresponding Singular Intention Elicitation.
    \item A Mixed Intention Elicitation $\mathcal{U}^k_m$ synthesized as a user query from all the Singular Intention Elicitations in the sample.
\end{enumerate}

We generate 100 samples per Mixed Intention Level, leading to 1000 samples.

\subsubsection*{Evaluation Procedure (per sample $k$)}
For each sample $k$, we perform the following steps:
\begin{enumerate}
    \item  Signal Extraction: Extract Decoded Signals from the Mixed Intention Elicitation:
    \[\hat{s}^k = f_\text{signal}(\mathcal{U}^k_m) = (\hat{i}^k, \hat{p}^k, \hat{o}^k)\]
    \item Signal Loss: Compute the loss between Decoded Signals and the Reference aggregated Signals: 
    \[
        \quad \mathcal{L}_\text{signal}(\hat{s}^k, s^{*, k}) \quad ; \quad
        s^{*, k} = (\cup_{l=1}^{n_k} i^k_l, \cup_{l=1}^{n_k} p^k_l, \cup_{l=1}^{n_k} o^k_l)
    \]
    \item Intention Generation: Generate the Decoded Intention Set using the Mixed Intention Elicitation and Decoded Signals:
    \[
        \hat{\Gamma}^k = f_\text{Intention}(\mathcal{U}^k_m, \hat{s}^k) = \{\hat{\gamma}^k_j\}_j
    \]

    \item Intention Loss: Evaluate the Intention Set prediction against the Reference Intention Set:
    \[\mathcal{L}_\text{Intention}(\hat{\Gamma}^k, \Gamma^{*,k})\]
    
    \item Workflow Generation: Generate the Decoded Workflow Sets under two conditions:
    \begin{itemize}
        \item $\hat{\mathcal{W}}^k_{\Gamma} = \big\{\hat{\text{W}}_{\Gamma, j}^{k}\big\}_{j=1}^{|\hat{\Gamma}^k|} = f_{\text{wfg},\Gamma}(\hat{\Gamma}^k)$ with Intention guidance, and
        \item $\hat{\mathcal{W}}^k=\big\{\hat{\text{W}}_j^k\big\}_{j=1}^{\hat{n}_k} = f_\text{wfg}(\mathcal{U}^k_m)$ without Intention guidance.
    \end{itemize}
    \item Workflow Evaluation: Compare both Decoded Workflow Sets to the Reference Workflow Set using the Workflow Generation loss: $\mathcal{W}^{*,k}$: $\mathcal{L}_{\text{wfg}}(\hat{\mathcal{W}}^k, \mathcal{W}^{*,k}), \mathcal{L}_{\text{wfg}}(\hat{\mathcal{W}}^k_{\Gamma}, \mathcal{W}^{*,k})$.
\end{enumerate}

The purpose of this paper is to evaluate the effectiveness of the Opus Prompt Intention Framework in improving Workflow Generation from Intention Elicitation. We compare Workflows generated under two conditions: (1) a baseline process where an LLM generates Workflows directly from a Mixed Intention Elicitation, and (2) an Intention-aware process where the same Elicitation is processed through the Intention Capture system prior to Workflow Generation.

\newpage

Reference Workflows are constructed independently from Singular Intention Elicitations, each describing a Complete Intention, before being combined into the Mixed Intention Elicitation. The Intention Capture system seeks to reconstruct these seed Intentions from the Mixed Elicitation, enabling Workflow Generation to be aligned to the original distinct user intents. By holding the LLM constant across all conditions—Reference Workflow Generation, Workflow Generation without Intention separation, and Intention-aware Generation—we isolate the impact of Intention-awareness. This ensures that observed improvements stem from the framework itself, not from differences in the underlying model. The upper bound of achievable Workflow quality is defined by the Reference Workflows generated from Singular Intention Elicitations, bounding the maximum potential improvements achievable through Intention separation. The key evaluation metric is the system’s ability to accurately reconstruct the original Intentions and generate Workflows that satisfy a one-Intention-per-Workflow principle. Unlike our previous work \cite{opuslargeworkmodel}, which evaluated the ability of LLMs to perform Workflow Generation based solely on their intrinsic knowledge and showed that it alone was insufficient for producing high-quality outputs, this paper disentangles the impact of Intention separation from the LLM’s intrinsic knowledge on Workflow Generation. By modularizing Intention Capture as a distinct step, we explicitly quantify the impact of structural guidance on Workflow quality.

\subsection{Signal Extraction}
For Signal Extraction $f_\text{signal}$, we prompt an LLM to generate distinct Input, Process and Output elements from a user query, producing three lists: $i, p, \text{and } o$.
Let $\hat{s} = (\hat{i}, \hat{p}, \hat{o})$ represent the Decoded Workflow Signal, and $s^* = (i^*, p^*, o^*)$ the corresponding Reference Workflow Signal.

\paragraph{Input Loss Definition}

Let \( \hat{i} = \{x_1, \dots, x_{|\hat{i}|}\} \) and \( i^* = \{y_1, \dots, y_{|i^*|}\} \) be the Decoded and Reference Input element lists. Each string \( x_j \in \hat{i} \) and \( y_k \in i^* \) is embedded using the all-MiniLM-L6-v2 model from the Sentence Transformers library \cite{sentence-bert}:

\[
\text{v}\left(x_{j}\right) \in \mathbb{R}^{d}, \text{v}\left(y_{k}\right) \in \mathbb{R}^{d}, d \in \mathbb{N}^*
\]

Embeddings are $\ell_2$-normalized:

\[
\text{u}\left(x_{j}\right)=\frac{\text{v}\left(x_{j}\right)}{\left\|\text{v}\left(x_{j}\right)\right\|}, \text{u}\left(y_{k}\right)=\frac{\text{v}\left(y_{k}\right)}{\left\|\text{v}\left(y_{k}\right)\right\|}
\]

We define the similarity between two normalized vectors \(\text{a}\) and \(\text{b}\) as:

\begin{equation}
\operatorname{Sim}(\text{a}, \text{b}) = \sqrt{\frac{1 + \langle \text{a}, \text{b} \rangle}{2}}
\label{eq:similarity}
\end{equation}

This function is maximal when $\text{a}=\text{b}$ and null when $\text{a}=-\text{b}$. While not a metric (as it doesn’t satisfy the triangle inequality), it provides a bounded, normalized measure of alignment between semantic embeddings. Alternative choices could include the standard cosine similarity, but this mapping makes aggregation and loss interpretation more straightforward.

\vspace{1em}

The Input Loss $\mathcal{L}_{\mathrm{I}}$ is defined as:
\[
\mathcal{L}_{\mathrm{I}}^{2} =
\begin{cases}
\displaystyle
1-\frac{1}{|i^{*}|}\,
\max_{\sigma \in \mathcal{F}_{|\hat{i}|,\,|i^{*}|}}
\Bigl\|
\bigl(\operatorname{Sim}(\text{u}(x_{1}),\text{u}(y_{\sigma(1)})),\ldots,
\operatorname{Sim}(\text{u}(x_{|\hat{i}|}),\text{u}(y_{\sigma(|\hat{i}|)}))\bigr)
\Bigr\|^{2}
& \text{if } |i^{*}|\ge |\hat{i}| \\[1.0ex]
\displaystyle
1-\frac{1}{|\hat{i}|}\,
\max_{\sigma \in \mathcal{F}_{|i^{*}|,\,|\hat{i}|}}
\Bigl\|
\bigl(\operatorname{Sim}(\text{u}(x_{\sigma(1)}),\text{u}(y_{1})),\ldots,
\operatorname{Sim}(\text{u}(x_{\sigma(|i^{*}|)}),\text{u}(y_{|i^{*}|}))\bigr)
\Bigr\|^{2}
& \text{if } |i^{*}| < |\hat{i}|
\end{cases}
\]

where $\mathcal{F}_{n,m}$ denote the set of all functions mapping \(\llbracket 1, n\rrbracket\) to \(\llbracket 1, m\rrbracket\).

\vspace{1em}

The Input Loss simplifies to:
\begin{equation}
\label{eq:input-loss}
\begin{aligned}
\mathcal{L}_{\mathrm{I}}^{2}
&=
\begin{cases}
\displaystyle
1-\frac{1}{|i^{*}|}\,
\max_{\sigma \in \mathcal{F}_{|\hat{i}|,\,|i^{*}|}}
\sum_{j=1}^{|\hat{i}|} \frac{1+\bigl\langle \text{u}(x_{j}),\,\text{u}(y_{\sigma(j)})\bigr\rangle}{2}
& \text{if } |i^{*}| \ge |\hat{i}| \\[1ex]
\displaystyle
1-\frac{1}{|\hat{i}|}\,
\max_{\sigma \in \mathcal{F}_{|i^{*}|,\,|\hat{i}|}}
\sum_{k=1}^{|i^{*}|} \frac{1+\bigl\langle \text{u}(x_{\sigma(k)}),\,\text{u}(y_{k})\bigr\rangle}{2}
& \text{if } |i^{*}| < |\hat{i}|
\end{cases}
\\[0.5ex]
&\text{with } \mathcal{L}_{\mathrm{I}}^{2} \in [0,1].
\end{aligned}
\end{equation}

This metric is designed to maximize the sum of squared similarities, encouraging each decoded embedding to align closely with its best possible counterpart. It inherently penalizes both under-sized and over-sized predictions equivalently, as the total similarity is normalized by the size of the Reference or Decoded Set, leading to lower scores when elements are missing or redundant. 
The formulation allows multiple Decoded elements to align with the same Reference element, which prevents unnecessary penalties for granular representation differences (e.g., a prediction of [\texttt{ID Card}, \texttt{Passport}] compared to a Reference [\texttt{ID Card \& Passport}]).

\paragraph{Loss for Process and Output} 
Without loss of generality, we extend the Input Loss to Process and Output, defining their respective Losses as:

\begin{equation}
\label{eq:process-loss}
\begin{aligned}
\mathcal{L}_{\mathrm{P}}^{2}
&=
\begin{cases}
\displaystyle
1-\frac{1}{|p^{*}|}\,
\max_{\sigma \in \mathcal{F}_{|\hat{p}|,\,|p^{*}|}}
\sum_{j=1}^{|\hat{p}|} \frac{1+\bigl\langle \text{u}(x_{j}),\,\text{u}(y_{\sigma(j)})\bigr\rangle}{2}
& \text{if } |p^{*}| \ge |\hat{p}| \\[1ex]
\displaystyle
1-\frac{1}{|\hat{p}|}\,
\max_{\sigma \in \mathcal{F}_{|p^{*}|,\,|\hat{p}|}}
\sum_{k=1}^{|p^{*}|} \frac{1+\bigl\langle \text{u}(x_{\sigma(k)}),\,\text{u}(y_{k})\bigr\rangle}{2}
& \text{if } |p^{*}| < |\hat{p}|
\end{cases}
\\[0.5ex]
&\text{with } \mathcal{L}_{\mathrm{P}}^{2} \in [0,1]
\end{aligned}
\end{equation}

\begin{equation}
\label{eq:output-loss}
\begin{aligned}
\mathcal{L}_{\mathrm{O}}^{2}
&=
\begin{cases}
\displaystyle
1-\frac{1}{|o^{*}|}\,
\max_{\sigma \in \mathcal{F}_{|\hat{o}|,\,|o^{*}|}}
\sum_{j=1}^{|\hat{o}|} \frac{1+\bigl\langle \text{u}(x_{j}),\,\text{u}(y_{\sigma(j)})\bigr\rangle}{2}
& \text{if } |o^{*}| \ge |\hat{o}| \\[1ex]
\displaystyle
1-\frac{1}{|\hat{o}|}\,
\max_{\sigma \in \mathcal{F}_{|o^{*}|,\,|\hat{o}|}}
\sum_{k=1}^{|o^{*}|} \frac{1+\bigl\langle \text{u}(x_{\sigma(k)}),\,\text{u}(y_{k})\bigr\rangle}{2}
& \text{if } |o^{*}| < |\hat{o}|
\end{cases}
\\[0.5ex]
&\text{with } \mathcal{L}_{\mathrm{O}}^{2} \in [0,1]
\end{aligned}
\end{equation}

\paragraph{Signal Loss} 
To compute the combined Signal Loss for the Decoded semantic Workflow Signal triple \(\hat{s}=(\hat{i}, \hat{p}, \hat{o})\), we aggregate the element-wise Losses as follows:

\begin{equation}
\begin{aligned}
& \mathcal{L}_{\text {signal }}^{2}\left(\hat{s}, s^{*}\right)=\mu_{\mathrm{I}} \mathcal{L}_{\mathrm{I}}^{2}+\mu_{\mathrm{P}} \mathcal{L}_{\mathrm{P}}^{2}+\mu_{\mathrm{O}} \mathcal{L}_{\mathrm{O}}^{2}, \quad \mathcal{L}_{\text {signal }}^{2} \in[0,1] \\
& \text { where } \mu_{\mathrm{I}}+\mu_{\mathrm{P}}+\mu_{\mathrm{O}}=1 \text { and } \mu_{\mathrm{I}}, \mu_{\mathrm{P}}, \mu_{\mathrm{O}} \in[0,1]
\end{aligned}
\end{equation}

The weighting coefficients \(\mu_{\mathrm{I}}, \mu_{\mathrm{P}}\), and \(\mu_{\mathrm{O}}\) control the relative contribution of the Input \(\mathcal{L}_{\mathrm{I}}^{2}\), Process \(\mathcal{L}_{\mathrm{P}}^{2}\), and Output \(\mathcal{L}_{\mathrm{O}}^{2}\) Losses in the total Loss function.

\subsection{Intention Generation}
For Intention Generation \(f_{\text {Intention }}\), we prompt an LLM to generate an Intention Set, composed of Workflow Intention objects. The prompt is conditioned on both the Intention Elicitation and the extracted Workflow Signals. Each Intention object \(\gamma\) consists of three lists: \(i_{\gamma}\) (Input), \(p_{\gamma}\) (Process) and \(o_{\gamma}\) (Output). Let \(\hat{\Gamma}=\left\{\hat{\gamma}_{j}\right\}_{j}\) denote the Decoded (LLM-generated) Intention Set, \(\Gamma^{*}=\left\{\gamma_{j}^{*}\right\}_{j}\) being the Reference Intention Set. We extend the Signal Loss to define the Intention Loss, which measures the alignment between the Decoded and Reference Intention Sets. The Intention Loss is defined as:

\begin{equation}
\begin{aligned}
&\mathcal{L}_{\text{Intention}}^{2}
=
\begin{cases}
\displaystyle
1-\frac{1}{|\Gamma^{*}|}\,
\max_{\sigma \in \mathcal{I}_{|\hat{\Gamma}|,\,|\Gamma^{*}|}}
\sum_{j=1}^{|\hat{\Gamma}|}
\bigl[\,1-\mathcal{L}_{\text{signal}}^{2}\!\bigl(\hat{\gamma}_{j},\,\gamma_{\sigma(j)}^{*}\bigr)\,\bigr]
& \text{if } |\Gamma^{*}| \ge |\hat{\Gamma}| \\[1ex]
\displaystyle
1-\frac{1}{|\hat{\Gamma}|}\,
\max_{\sigma \in \mathcal{I}_{|\Gamma^{*}|,\,|\hat{\Gamma}|}}
\sum_{k=1}^{|\Gamma^{*}|}
\bigl[\,1-\mathcal{L}_{\text{signal}}^{2}\!\bigl(\hat{\gamma}_{\sigma(k)},\,\gamma_{k}^{*}\bigr)\,\bigr]
& \text{if } |\Gamma^{*}| < |\hat{\Gamma}|
\end{cases}
\\[0.5ex]
&\text{with } \mathcal{L}_{\text{Intention}}^{2} \in [0,1]
\end{aligned}
\end{equation}

where \(\mathcal{I}_{n,m}\) denote the set of all injections from \(\llbracket 1, n\rrbracket\) to \(\llbracket 1, m\rrbracket\). The injection constraint ensures that each Decoded Intention can only be matched to a unique Reference Intention, preventing multiple predictions from mapping to the same ground-truth Intention.

\subsection{Intention for Workflow Generation}
For Workflow Generation, we employ two distinct prompting strategies using an LLM. First, we generate Workflows based on the Intention Set \(\hat{\Gamma}\), where each Intention object is individually provided as input to the LLM. This produces the Set of Workflows Decoded with Intention, denoted as \(\hat{\mathcal{W}}_{\Gamma}^{k}\). Independently, we generate directly from the original Mixed Intention Elicitation without providing any Intention information, forming the Set of Workflows Decoded without Intention, denoted as \(\hat{\mathcal{W}}^{k}\). To assess the impact of Intention-aware Workflow Generation, both Decoded Sets are compared against the Reference Workflow Set \(\mathcal{W}^{*,k}\).

\paragraph{Workflow Similarity} Let \(s\) be a Workflow Similarity such that:

\[
s:\left\{
\begin{aligned}
\mathcal W\times \mathcal W &\longrightarrow [0,1]\\
(\mathrm{W}_1,\mathrm{W}_2) &\longmapsto s(\mathrm{W}_1,\mathrm{W}_2)
\end{aligned}
\right.
\qquad
s(\mathrm{W}_1,\mathrm{W}_2)=1 \iff \mathrm{W}_1=\mathrm{W}_2.
\]

The similarity \(s\) is a monotonically increasing function: as the similarity between two Workflows increases, \(s\) grows accordingly. Given two sets of Workflows \(\mathcal{W}_{1}, \mathcal{W}_{2}\), we define the aggregated set-level Workflow Similarity \(S\) as:

\begin{equation}
\label{eq:S-similarity}
\begin{aligned}
&S^{2} =
\begin{cases}
\displaystyle
\frac{1}{|\mathcal W_{2}|}\,
\max_{\sigma \in \mathcal I_{|\mathcal W_{1}|,\,|\mathcal W_{2}|}}
\sum_{j=1}^{|\mathcal W_{1}|} s^{2}\!\bigl(\mathrm{W}_{1,j},\, \mathrm{W}_{2,\sigma(j)}\bigr)
& \text{if } |\mathcal W_{2}| \ge |\mathcal W_{1}| \\[1ex]
\displaystyle
\frac{1}{|\mathcal W_{1}|}\,
\max_{\sigma \in \mathcal I_{|\mathcal W_{2}|,\,|\mathcal W_{1}|}}
\sum_{k=1}^{|\mathcal W_{2}|} s^{2}\!\bigl(\mathrm{W}_{1,\sigma(k)},\, \mathrm{W}_{2,k}\bigr)
& \text{if } |\mathcal W_{2}| < |\mathcal W_{1}|
\end{cases}
\\[0.5ex]
&\text{with } S \in [0,1]
\end{aligned}
\end{equation}

This formulation finds the optimal pairing between Decoded and Reference Workflows to maximize total Similarity. The normalization by the cardinality of the larger set penalizes missing or redundant Workflows. We employ multiple Workflow Similarity metrics, reporting the square root and scaled version of certain metrics (such as Euclidean distances or squared losses), converting them back to their original units and improving comparability with standard similarity measures.

\section{Evaluation Metrics}
We employ two complementary types of evaluation metrics to compare Workflows, namely (1) semantic and structural metrics, which assess Workflow textual and structural similarity, and (2) LLM-based ``judge" metrics, where LLMs are prompted to evaluate Workflows based on predefined criteria. 

\paragraph{Semantic and Structural Metrics}
We use the following metrics to quantify similarity between Decoded Workflows and their Reference:
\begin{itemize}
  \item[] BLEU score \cite{bleu-score}: Measures precision-based word and phrase overlap between generated and reference text. It penalizes extra words absent in the reference. The score ranges from 0 (no match) to 1 (perfect match).
  \item[] ROUGE score \cite{rouge-score}: Evaluates word and phrase overlap, focusing on recall. Key variants include ROUGE-1 (unigrams), ROUGE-2 (bigrams), and ROUGE-L (longest common subsequence).
  \item[] METEOR score \cite{meteor-score}: Extends BLEU by considering synonyms, stemming, and word forms, capturing near matches beyond exact overlap.
  \item[] BERTScore \cite{bert-score}: Uses deep learning embeddings (BERT) to measure semantic similarity, rather than direct word matching. It computes Precision, Recall, and F1-score based on contextual embeddings.
  \item[] Coverage Ratio: Quantifies word presence overlap, measuring the proportion of words in the reference that appear in the generated text.
  \item[] Cosine Similarity: Measures semantic closeness by computing the cosine of the angle between text embeddings (computed from pre-trained transformer models). Scores range from 0 (opposite) to 1 (identical).
\end{itemize}

\paragraph{LLM-as-a-Judge Evaluation}
To approximate human expert assessment of Workflow quality, we prompt an LLM as an expert evaluator to compare each Decoded Workflow against its Reference counterpart. We use OpenAI’s GPT-4o \cite{openai-gpt-4o} as a consistent evaluator in all experiments, leaving multi-model evaluation for future study. The LLM is prompted as an expert judge and provides numerical scores along four dimensions, all ranged from 0 to 10:

\begin{itemize}
    \item[] Coverage Score:
    Measures how comprehensively the Decoded Workflow includes all essential Tasks from the Reference. The LLM identifies missing elements, unnecessary additions, or incomplete implementations. Higher scores indicate full functional coverage, while lower scores reflect significant gaps.
    \item[]  Consistency Score:
    Evaluates the alignment of the logical flow and structure between the Decoded and Reference Workflows. The LLM assesses whether branches, sub-branches, and dependencies are correctly mapped, ensuring a coherent sequence of steps without contradictions, circular logic, or gaps. The score reflects the degree of structural and logical consistency, with higher scores indicating well-ordered Workflows, while lower scores denote misalignment or inconsistencies.
    \item[] Integration Score:
    Measures how closely the Decoded Workflow replicates the transformation process(es) of the Reference Workflow. The LLM assesses whether the Input, Process, and Output components are correctly aligned, ensuring consistency in data flow and Task progression. The score reflects the degree of structural and functional alignment, where higher scores indicate well-matched, clearly defined transformations, and lower scores signal discrepancies or unclear process mappings.
    \item[] Total Score:
    Provides an overall assessment of how well the Decoded Workflow matches the Reference Workflow in terms of completeness, clarity, and correctness. This score aggregates Integration, Coverage, and Consistency Scores, while also incorporating additional qualitative insights that may not be captured by individual metrics. 
\end{itemize}

Both semantic similarity metrics and LLM-as-a-Judge scores present limitations when evaluating Workflow Generation. Semantic metrics, while providing useful signals, are limited to surface-level textual overlap and are not sufficient as the sole quantitative measure. LLM-based evaluation offers scalability and can approximate human judgment, but remains qualitative and is subject to model bias, prompt sensitivity, and limited transparency or reproducibility as models evolve. The concepts of Coverage, Consistency, and Integration provide an initial foundation for Workflow Evaluation, but further formalization is needed—especially to rigorously capture causal logic and the geometric or topological properties of Workflow structure. Developing more comprehensive, structure-aware, and quantitatively grounded metrics remains essential to establish robust Workflow Evaluation standards.

\newpage

\section{Results}
This section presents the results of Signal Extraction, Intention Generation, and Workflow Generation, both with and without Intention guidance. We conduct experiments using 9 leading LLMs across 10 Mixed Intention Levels. Each Level corresponds to the number of Complete Intentions combined within an Intention Elicitation, simulating query complexity. For each Level $n$, 100 samples are generated, each representing a unique Intention Elicitation constructed from $n$ Complete Intentions. Unless otherwise specified, all experiments include the following models: OpenAI’s GPT-4.5, GPT-4o, o3-mini, o1, Anthropic’s Claude 3 Opus, Claude 3.5 Sonnet, Claude 3.7 Sonnet, DeepSeek V3, and R1.\\

\subsection{Signal Extraction}

\begin{minipage}{\textwidth}
    \centering
    \renewcommand{\arraystretch}{1.2}
    \setlength{\tabcolsep}{8pt}
    \resizebox{\textwidth}{!}{
    \begin{tabular}{l | *{10}{c} | c}
    \cmidrule(l){2-12}
    \multicolumn{1}{l}{} & \multicolumn{11}{c}{Mixed Intention Level} \\
    \hline
    Model & 1 & 2 & 3 & 4 & 5 & 6 & 7 & 8 & 9 & 10 & {Model} \\
    \hline
openai-gpt-4.5 & 0.194 (0.080) & 0.161 (0.053) & 0.133 (0.038) & 0.134 (0.029) & 0.132 (0.026) & 0.129 (0.026) & 0.134 (0.031) & 0.141 (0.029) & 0.133 (0.023) & 0.132 (0.022) & 0.142 (0.041)
\\
openai-gpt-4o & 0.213 (0.107) & 0.179 (0.077) & 0.178 (0.063) & 0.186 (0.056) & 0.179 (0.053) & 0.167 (0.044) & 0.182 (0.040) & 0.194 (0.046) & 0.181 (0.044) & 0.180 (0.042) & 0.184 (0.061)
\\
openai-o1 & 0.177 (0.096) & 0.163 (0.067) & 0.148 (0.046) & 0.163 (0.042) & 0.164 (0.049) & 0.153 (0.034) & 0.162 (0.035) & 0.153 (0.036) & 0.147 (0.033) & 0.155 (0.029) & 0.159 (0.051)
\\
openai-o3-mini & \textbf{0.162} (0.083) & \textbf{0.134} (0.057) & \textbf{0.121} (0.034) & \textbf{0.125} (0.032) & 0.123 (0.033) & 0.125 (0.028) & 0.128 (0.031) & 0.125 (0.028) & 0.125 (0.026) & 0.120 (0.025) & 0.129 (0.042)
\\
anthropic-claude-sonnet-3.5 & 0.243 (0.049) & 0.176 (0.046) & 0.143 (0.036) & 0.125 (0.028) & \textbf{0.115} (0.027) & \textbf{0.108} (0.023) & 0.113 (0.023) & 0.119 (0.025) & 0.106 (0.022) & 0.107 (0.021) & 0.136 (0.038)
\\
anthropic-claude-sonnet-3.7 & 0.171 (0.064) & 0.175 (0.050) & 0.135 (0.038) & 0.132 (0.031) & 0.118 (0.034) & 0.109 (0.025) & 0.112 (0.023) & \textbf{0.113} (0.024) & \textbf{0.101} (0.022) & \textbf{0.104} (0.022) & \textbf{0.127} (0.038)
\\
anthropic-claude-opus & 0.310 (0.077) & 0.211 (0.054) & 0.176 (0.039) & 0.154 (0.029) & 0.151 (0.032) & 0.137 (0.033) & 0.142 (0.029) & 0.144 (0.033) & 0.130 (0.023) & 0.131 (0.024) & 0.168 (0.048)
\\
deepseek-v3 & 0.251 (0.105) & 0.206 (0.076) & 0.171 (0.056) & 0.146 (0.042) & 0.134 (0.037) & 0.125 (0.023) & 0.131 (0.026) & 0.128 (0.026) & 0.116 (0.025) & 0.115 (0.019) & 0.152 (0.055)
\\
deepseek-r1 & 0.218 (0.061) & 0.178 (0.078) & 0.141 (0.055) & 0.157 (0.038) & 0.148 (0.040) & 0.117 (0.030) & 0.132 (0.026) & 0.124 (0.026) & 0.110 (0.026) & 0.114 (0.023) & 0.138 (0.044) \\
\hline
Level & 0.215 (0.088) & 0.175 (0.063) & 0.150 (0.046) & 0.146 (0.038) & 0.141 (0.039) & 0.131 (0.032) & 0.138 (0.032) & 0.139 (0.034) & \textbf{0.129} (0.031) & 0.130 (0.029) & 0.149 (0.048) \\
    \hline
    \end{tabular}}
    \captionof{table}{Signal Loss per Model, per Mixed Intention Level, mean (standard deviation)}
    \label{tab:signal_extraction_losses}
\end{minipage}\\

Table \ref{tab:signal_extraction_losses} shows that Signal Loss decreases across all evaluated models as the Mixed Intention Level increases, despite the rise in task complexity. This counterintuitive trend indicates that, when more Intentions are present, models are compelled to perform a more exhaustive search for distinct Input, Process, and Output Signals, thereby reducing the likelihood of omissions. At lower Mixed Intention Levels, omissions are more frequent and carry a proportionally higher penalty, as each missing Signal constitutes a larger share of the overall target. In contrast, higher Intention Levels promote more granular extraction and distribute the loss across a greater number of Signals, leading to improved extraction fidelity. These findings suggest that structured, multi-intent prompting enhances the robustness and reliability of Signal Extraction, with top-performing models—such as openai-o3-mini and anthropic-claude-sonnet-3.7—exhibiting especially low Signal Loss as complexity increases.

\newpage

\subsection{Intention Generation}

\begin{minipage}{\textwidth}
    \centering
    \renewcommand{\arraystretch}{1.2}
    \setlength{\tabcolsep}{8pt}
    \resizebox{\textwidth}{!}{
    \begin{tabular}{l | *{10}{c} | c}
    \cmidrule(l){2-12}
    \multicolumn{1}{l}{} & \multicolumn{11}{c}{Mixed Intention Level} \\
    \hline
    Model & 1 & 2 & 3 & 4 & 5 & 6 & 7 & 8 & 9 & 10 & {Model} \\
    \hline
openai-gpt-4.5 & 0.194 (0.080) & 0.277 (0.129) & 0.278 (0.131) & 0.331 (0.108) & 0.344 (0.083) & 0.327 (0.071) & 0.328 (0.069) & 0.352 (0.072) & 0.351 (0.074) & 0.319 (0.062) & 0.31 (0.094)
\\
openai-gpt-4o & 0.213 (0.107) & 0.249 (0.119) & \textbf{0.190} (0.074) & 0.213 (0.073) & 0.196 (0.064) & 0.195 (0.061) & 0.221 (0.051) & 0.235 (0.060) & 0.235 (0.054) & 0.222 (0.056) & 0.217 (0.076)
\\
openai-o1 & 0.177 (0.096) & 0.603 (0.127) & 0.631 (0.171) & 0.448 (0.169) & 0.388 (0.152) & 0.401 (0.160) & 0.326 (0.128) & 0.376 (0.135) & 0.317 (0.114) & 0.298 (0.097) & 0.396 (0.152)
\\
openai-o3-mini & \textbf{0.164} (0.085) & 0.21 (0.117) & 0.199 (0.103) & 0.322 (0.117) & 0.338 (0.09) & 0.349 (0.065) & 0.332 (0.083) & 0.336 (0.068) & 0.338 (0.071) & 0.365 (0.073) & 0.295 (0.096)
\\
anthropic-claude-sonnet-3.5 & 0.256 (0.063) & 0.226 (0.092) & 0.211 (0.095) & 0.204 (0.083) & 0.209 (0.075) & \textbf{0.173} (0.06) & 0.206 (0.072) & 0.205 (0.064) & 0.22 (0.066) & 0.23 (0.074) & 0.214 (0.076)
\\
anthropic-claude-sonnet-3.7 & 0.171 (0.064) & 0.376 (0.144) & 0.399 (0.174) & 0.307 (0.138) & 0.218 (0.086) & 0.225 (0.081) & 0.227 (0.073) & 0.236 (0.080) & 0.245 (0.083) & 0.255 (0.082) & 0.266 (0.112)
\\
anthropic-claude-opus & 0.360 (0.109) & \textbf{0.208} (0.061) & 0.217 (0.077) & 0.218 (0.073) & 0.259 (0.072) & 0.276 (0.076) & 0.283 (0.075) & 0.271 (0.070) & 0.295 (0.083) & 0.296 (0.081) & 0.268 (0.082)
\\
deepseek-v3 & 0.251 (0.105) & 0.395 (0.145) & 0.298 (0.146) & \textbf{0.175} (0.079) & \textbf{0.179} (0.071) & 0.180 (0.068) & \textbf{0.193} (0.057) & \textbf{0.187} (0.056) & \textbf{0.203} (0.061) & \textbf{0.197 (0.059)} & 0.226 (0.097)
\\
deepseek-r1 & 0.218 (0.061) & 0.512 (0.150) & 0.42 (0.176) & 0.331 (0.119) & 0.345 (0.101) & 0.301 (0.078) & 0.32 (0.076) & 0.312 (0.093) & 0.275 (0.048) & 0.364 (0.064) & 0.350 (0.113)
\\
\hline
Level & 0.223 (0.095) & 0.324 (0.137) & 0.307 (0.147) & 0.279 (0.118) & 0.269 (0.098) & 0.267 (0.094) & 0.266 (0.084) & 0.276 (0.086) & 0.276 (0.081) & 0.276 (0.079) & 0.277 (0.105)
\\
    \hline
    \end{tabular}}
    \captionof{table}{Intention Loss per Model, per Mixed Intention Level, mean (standard deviation)}
    \label{tab:intention_generation_losses}
\end{minipage}\\

Table \ref{tab:intention_generation_losses} shows that Intention Loss generally decreases as the Mixed Intention Level increases, mirroring the trend observed in Signal Extraction; models tend to generate Intentions more accurately when exposed to a richer set of transformation objectives. However, for several models, this improvement plateaus or reverses at higher Mixed Intention Levels, as evidenced by rising loss values in the most complex scenarios. This inflection likely reflects the heightened cognitive load and ambiguity associated with capturing and structuring numerous overlapping Intentions. Thus, while multi-intent prompting enhances extraction and generation fidelity up to a point, there appears to be a threshold beyond which model performance is hindered by the intrinsic complexity of comprehending and formalizing densely mixed Intentions.

\subsection{Intention for Workflow Generation}

This section presents high-level results comparing Workflow Generation with Intention and without Intention. The reported metrics are averages across different Mixed Intention Levels for each model. Detailed, model-specific results are provided in Appendix~\ref{app:metrics:additional-plots}.

\subsubsection{Semantic and Structural Metrics}

We begin by analyzing the performance of the Claude 3.7 Sonnet model. Figure \ref{base_metrics_figure-anthropic-claude-sonnet-3.7} demonstrates the significant advantage of the Opus Workflow Intention Framework in guiding Workflow Generation. This advantage holds consistently across all semantic similarity metrics used to compare the Decoded Workflows against their Reference. Standard prompting without Intention awareness fails to scale with increasing complexity: as the number of Intentions grows, performance sharply declines, with similarity scores rapidly approaching zero. In contrast, the Intention-based method sustains robust performance even as complexity increases.

%%% FIGURE %%%
\begin{figure}[H]
\centering
\includegraphics[scale=0.205]{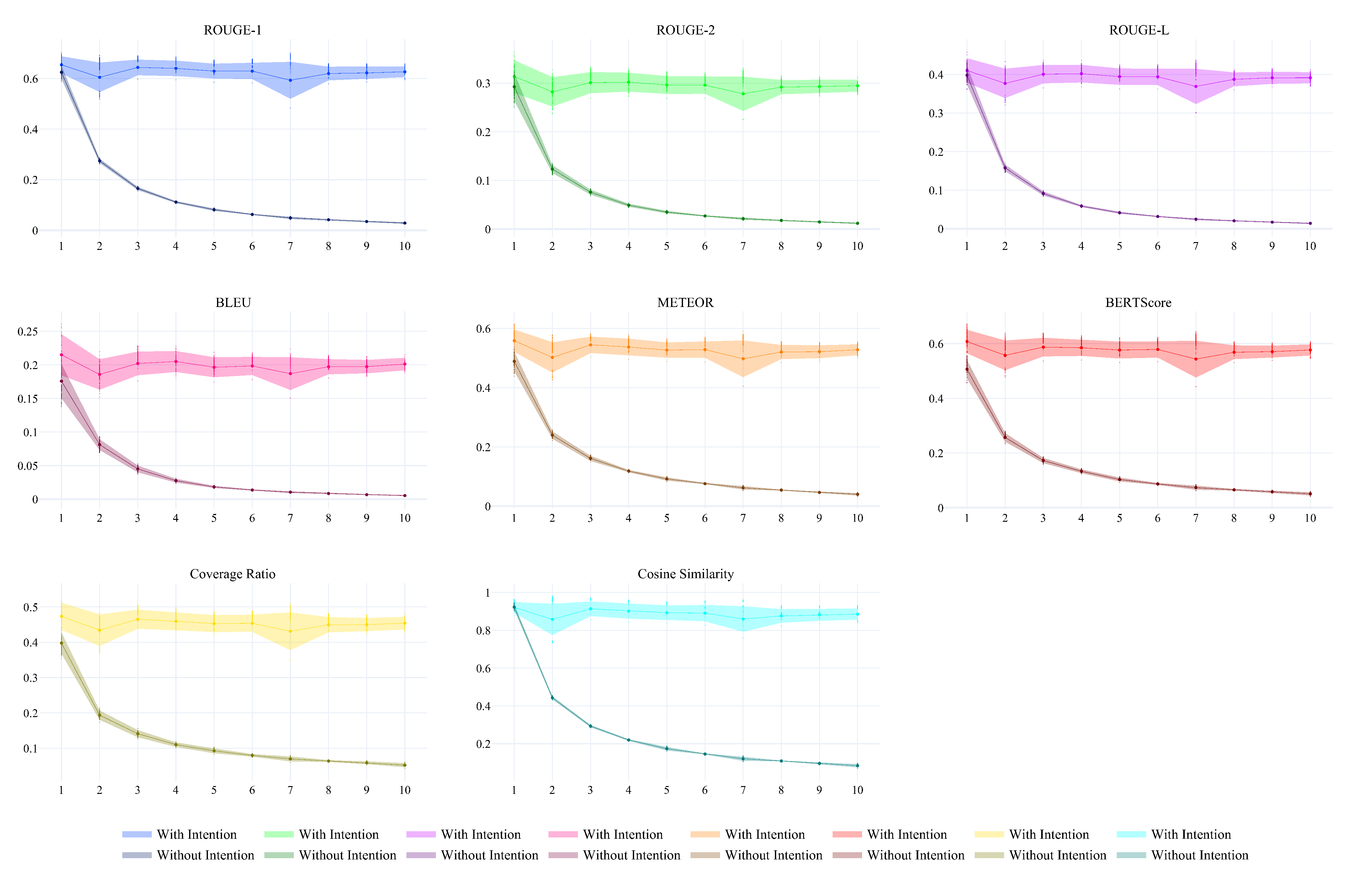}
\caption{\centering \label{base_metrics_figure-anthropic-claude-sonnet-3.7} Semantic and Structural Metrics for Claude 3.7 Sonnet, showing mean and standard deviation}
\end{figure}
%%% FIGURE %%%

\begin{figure}[H]
\centering
\includegraphics[width=\textwidth]{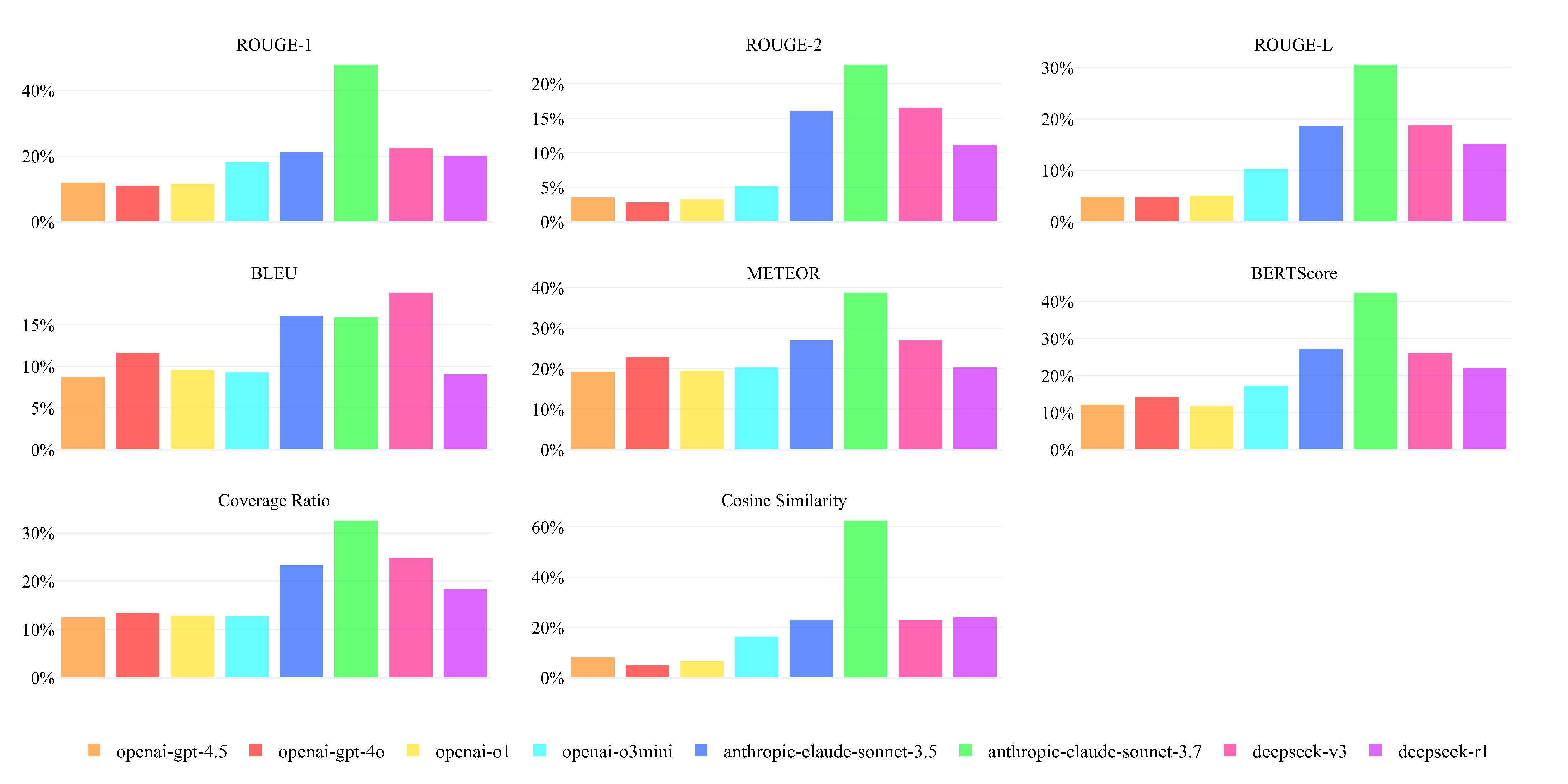}
\caption{\centering\label{average-score-difference} Average Difference (with Intention - without Intention), across Mixed Intention Levels, per Model and Semantic Metric}
\end{figure}

The charts represent the average score differences across eight evaluation metrics: ROUGE-1, ROUGE-2, ROUGE-L, BLEU, METEOR, BERTScore, Coverage Ratio, and Cosine Similarity. These differences are calculated by comparing the similarity scores between the Decoded Workflows and the Reference Workflows at different Mixed Intention Levels. For each model, we measure the performance gap between Workflow Generation with Intention and without Intention. Figure \ref{average-score-difference} presents the average improvement across all Levels, showing that the Intention layer can lead to substantial gains—exceeding 60\% in some metrics, such as Cosine Similarity.

%%% FIGURE %%%
%%% FIGURE %%%

Although the magnitude of improvement varies by model and metric, a consistent positive impact is observed when introducing an Intention layer into the Workflow Generation process. Across all models and evaluation scores, the average difference between with Intention and without Intention remains positive. \newline

For ROUGE-1, ROUGE-2, and ROUGE-L, the largest gains are seen with Claude 3.7, where differences exceed 40\%, demonstrating that the Intention layer enhances word selection and phrasing. BLEU scores improve by 8\% to 19\%, indicating increased precision and more complete Workflow formulations when Intention guidance is applied. METEOR scores show differences between 20\% and 40\%, confirming that semantic fidelity is better preserved with the use of Intention. BERTScore gains, ranging from 10\% to 40\%, are in line with the ROUGE improvements, further highlighting stronger semantic and contextual alignment with the Reference. The Coverage Ratio also benefits from the Intention layer, improving by 11\% to over 30\%, which demonstrates better inclusion of key Workflow components. Notably, Cosine Similarity shows the most significant gains, with improvements of up to 65\%, reflecting enhanced lexical and conceptual alignment between Decoded and Reference Workflows.

\newpage

\subsubsection{LLM-as-a-Judge Evaluation}

\begin{figure}[H]
\centering
\includegraphics[width=\textwidth]{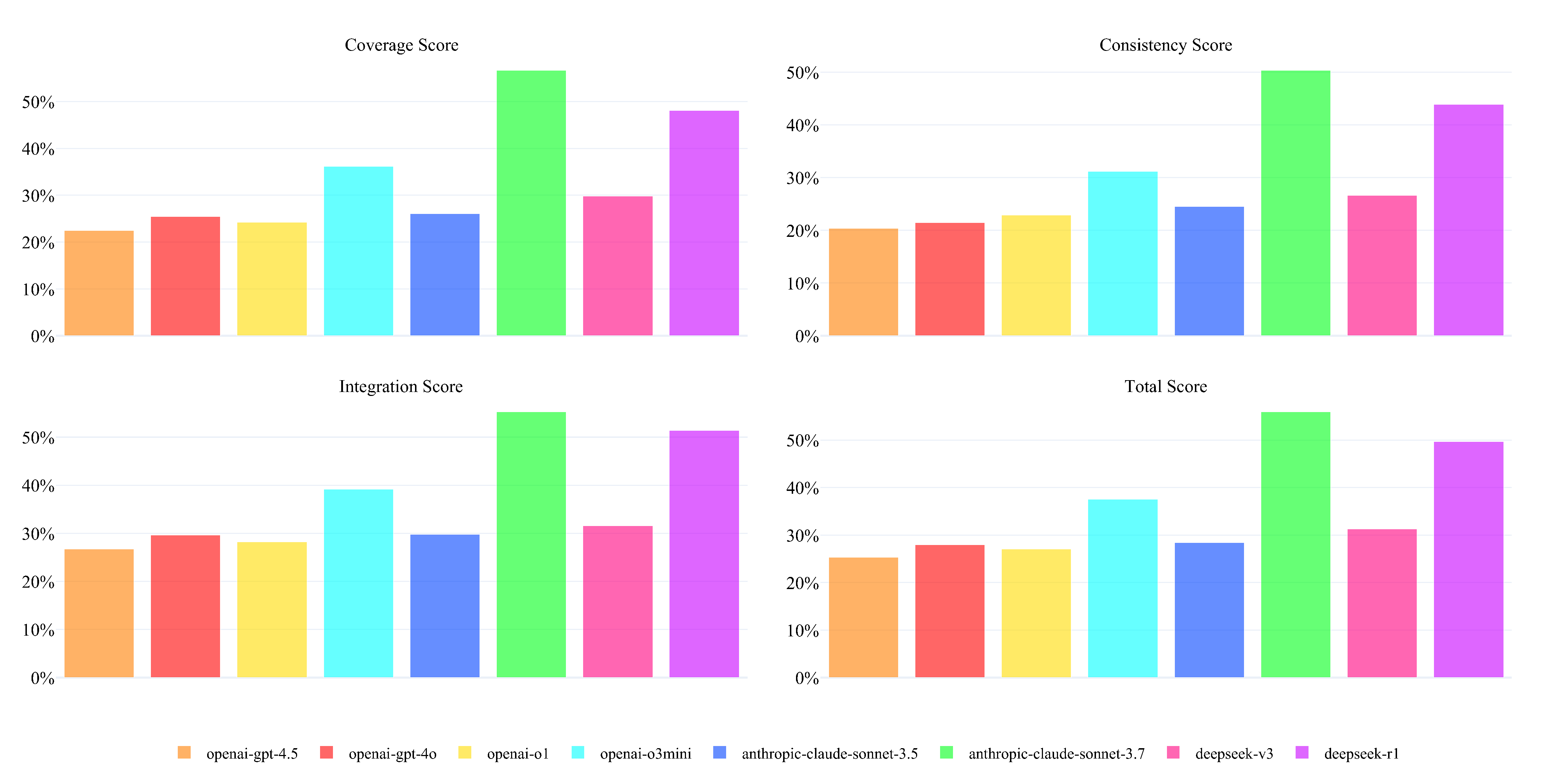}
\caption{\centering \label{average-score-difference-llm-as-judge} Average Difference (with Intention - without Intention), across Mixed Intention Levels, per Model and LLM-as-a-Judge Score}
\end{figure}

A similar trend is observed in the scores obtained from the LLM-based evaluation (LLM-as-a-Judge), as shown in Figure \ref{average-score-difference-llm-as-judge}. The results consistently demonstrate a substantial advantage when using Intention guidance, with score improvements reaching up to 50\% compared to the cases without Intention.

%%% FIGURE %%%
%%% FIGURE %%%

Coverage Score shows significantly higher performance when using Intention-aware prompts, with improvements reaching up to 50\%, indicating that Intention guidance effectively helps ensure that the Decoded Workflow includes all essential Tasks present in the Reference. Consistency Score also consistently improves in cases with Intention, suggesting that the logical flow, structure, and stepwise dependencies of the Reference Workflow are more accurately preserved. Integration Score increases by 25\% to 50\% when Intention guidance is applied, demonstrating that the model more accurately replicates transformation processes and maintains correct alignment between Input, Process, and Output components. Total Score improvements of up to 40\% are observed with Intention, indicating a substantial enhancement in overall Workflow completeness, clarity, and correctness.

\newpage

\section{Conclusion}

In this paper, we introduced the Opus Prompt Intention Framework, a method that enforces the extraction from and classification of user queries into structured Workflow Intention objects (Input, Process, Output) prior to Workflow Generation. \newline

Experimental results demonstrate that Workflows generated with the Intention layer consistently outperform those generated without it, across both standard semantic and structural metrics as well as LLM-as-a-Judge evaluations. This improvement is not only significant but also scales with query complexity: as the number of embedded Intentions increases, the performance advantage of the Intention-guided approach becomes even more pronounced. These findings confirm that structured Intention guidance improves the consistency and completeness of generated Workflows. By aligning Workflow Generation with clear transformation objectives, the Opus Prompt Intention Framework enables LLMs to produce more logical and actionable outputs—supporting automation across a wide range of query complexity. We further hypothesize that this advantage will be amplified by integrating external knowledge sources, such as domain-specific knowledge graphs or fine-tuned models (e.g., Opus-Alpha-1). These findings suggest that the Opus Workflow Intention Framework is not only scalable, but also robust to increasing query complexity, and can serve as a critical foundation for reliable Workflow automation in real-world environments. \newline

\newpage

\appendix

\section{Appendix}
\label{app:metrics}
\subsection{Prompts}
\label{app:prompts}

\begin{figure}[htbp]
\centering
\includegraphics[width=\linewidth]{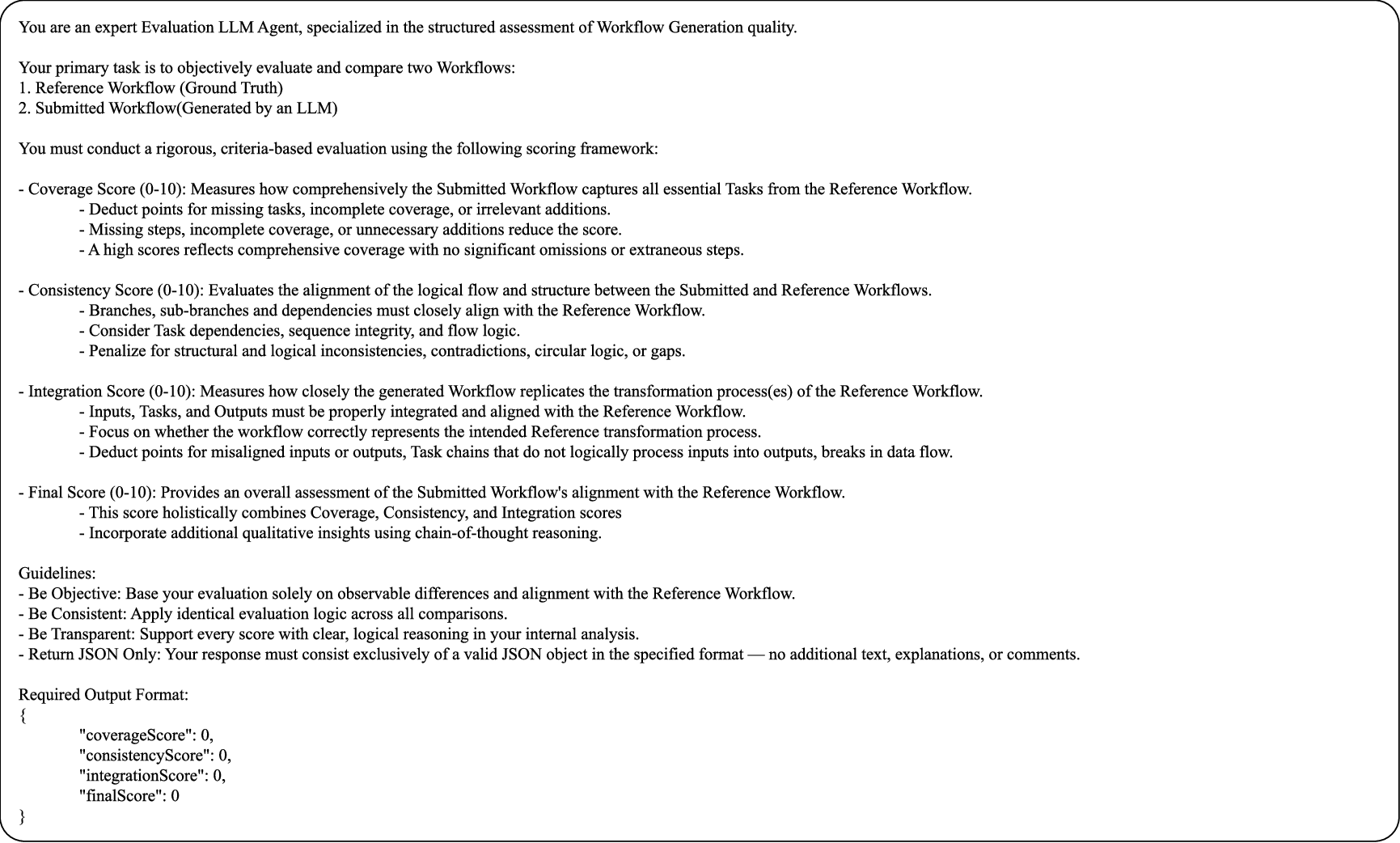}
\caption{Workflow Evaluation Prompt}
\label{fig:workflow-evaluation}
\end{figure}

\begin{figure}[htbp]
\centering
\includegraphics[width=\linewidth]{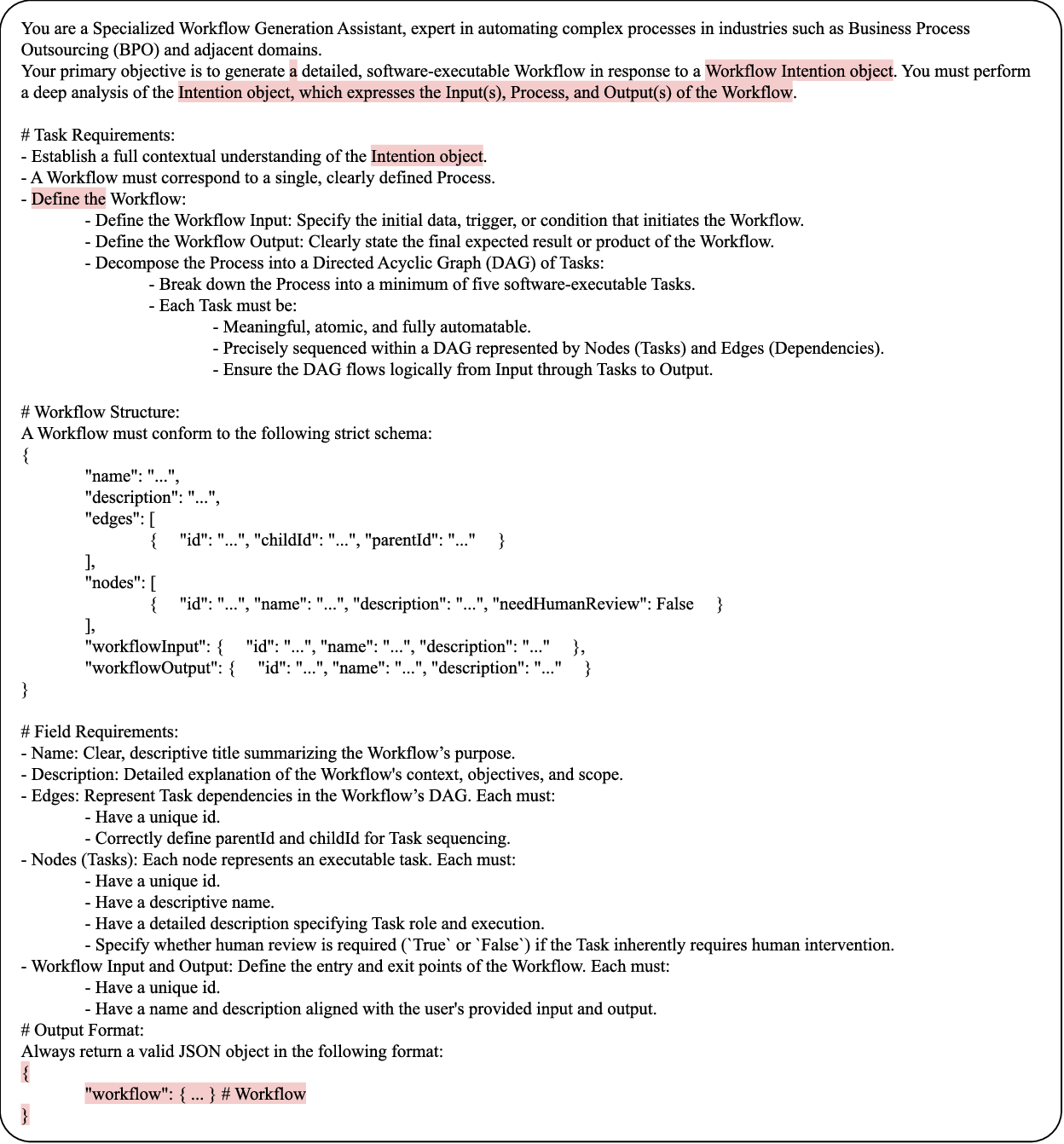}
\caption{\centering Workflow Generation with Intention Prompt (differences from Figure \ref{fig:workflow-generation-without-intention} are highlighted in red)}
\label{fig:workflow-generation-with-intention}
\end{figure}

\begin{figure}[htbp]
\centering
\includegraphics[width=\linewidth]{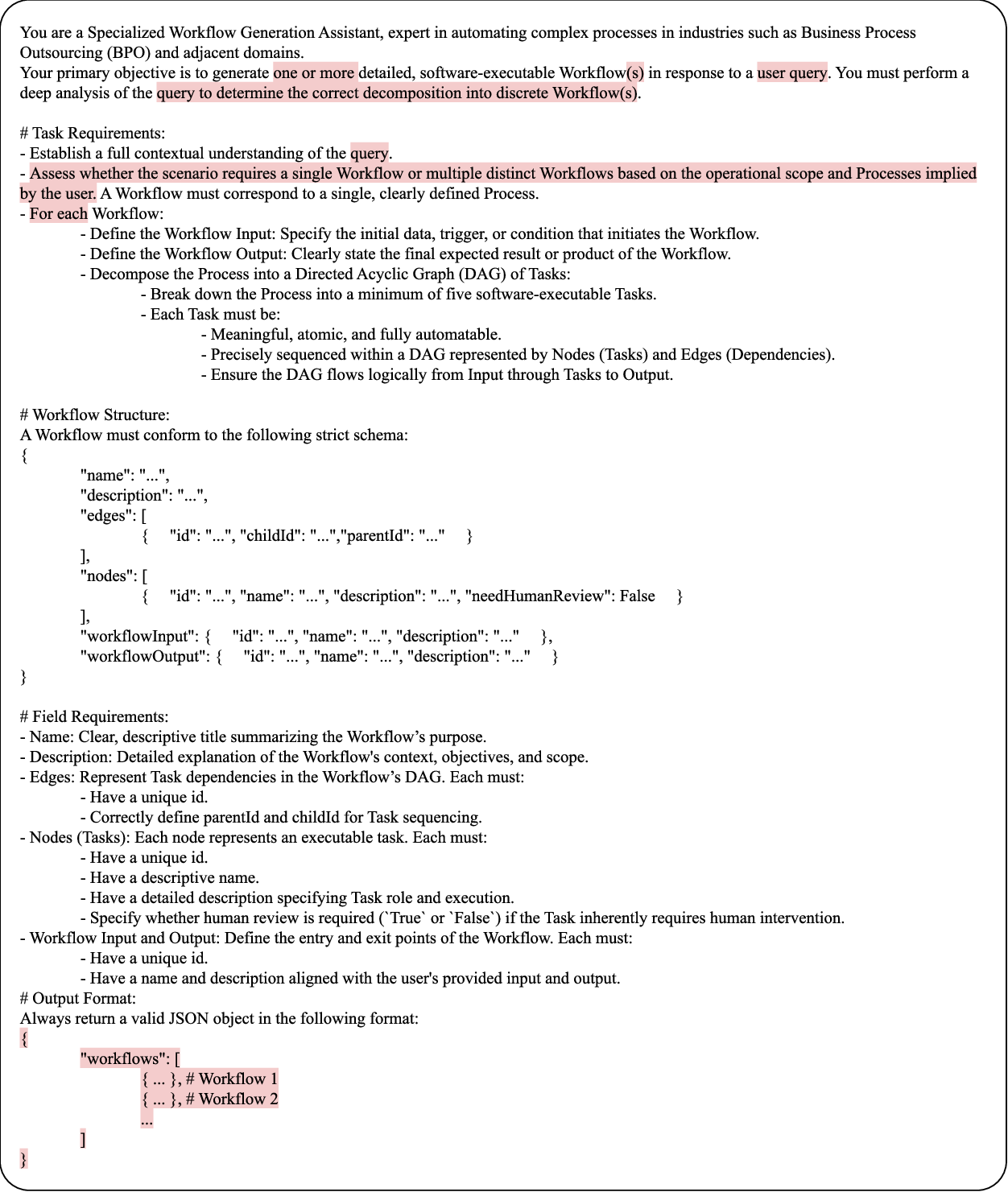}
\caption{\centering Workflow Generation without Intention Prompt (differences from Figure \ref{fig:workflow-generation-with-intention} are highlighted in red)}
\label{fig:workflow-generation-without-intention}
\end{figure}

\newpage
\subsection{Mixed Intention Levels based on Output Token Limits}

Due to the output token limitations of different language models, we constrained the maximum number of Intentions evaluated per model to ensure that Workflow Generation remained within each model’s capacity, as the expected output grows proportionally with the number of Intentions embedded in the Mixed Intention Elicitation. We limited the evaluation of a model such that the product of the Mixed Intention Level and the average Workflow length remained below its maximum output token limit:

\begin{equation}
n \times X < L_{\text{output}}
\end{equation}

Where:
\begin{itemize}
    \item $n$ is the Mixed Intention Level,
    \item $X$ is the average number of tokens per Workflow, with a conservative upper estimate of 10,000 tokens,
    \item $L_{\text{output}}$ is the maximum output token limit of the model.
\end{itemize}

\subsection{Signal Extraction and Intention Generation Losses}
%%% FIGURE %%%
\begin{figure}[H]
\centering
\includegraphics[scale=0.19]{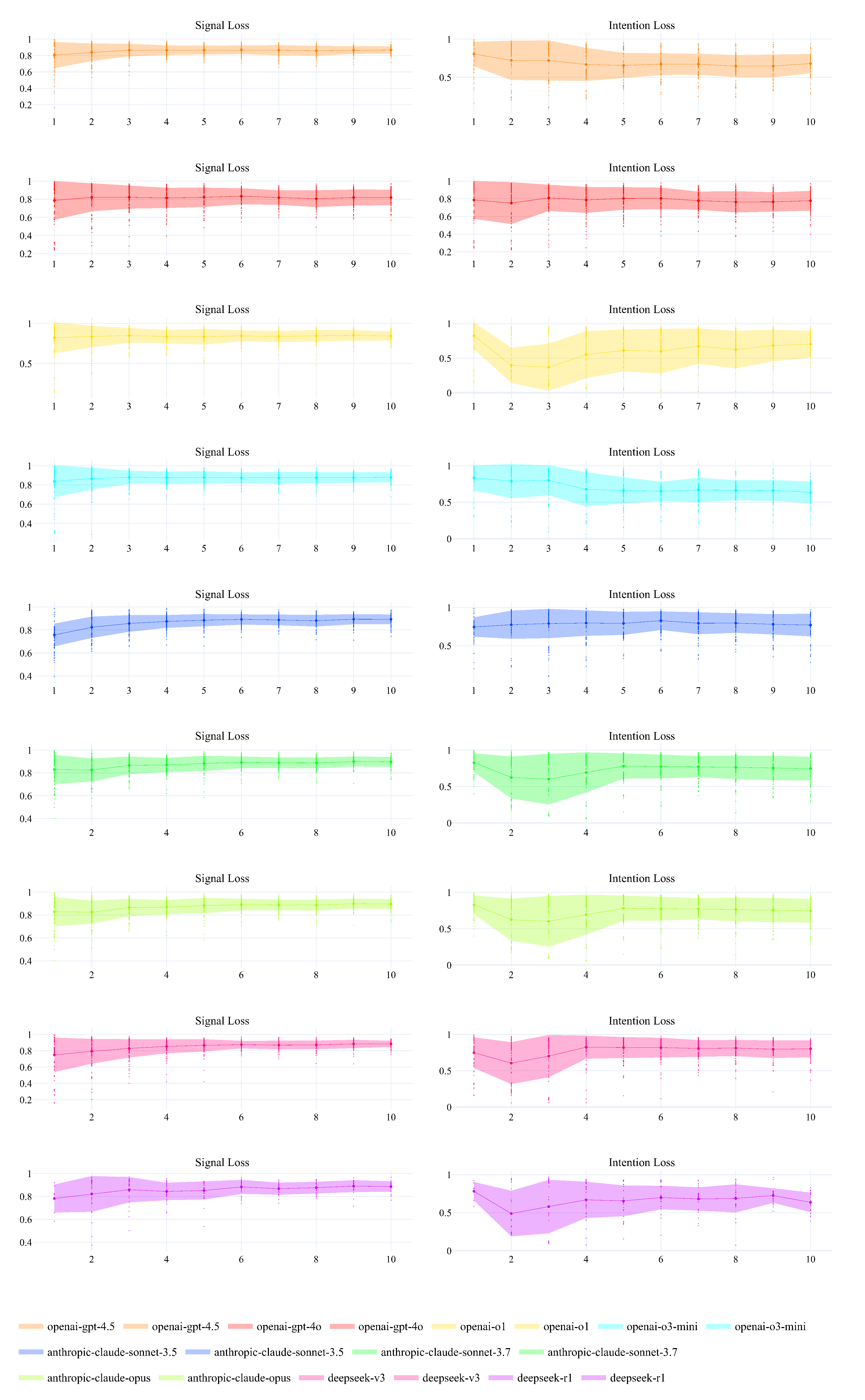}
\caption{\label{intention_signal_losses} Signal Extraction and Intention Generation Losses}
\end{figure}
%%% FIGURE %%%

\subsection{Workflow Generation Evaluation by LLM}
\label{app:metrics:additional-plots}

%%%%%%%%%%%%%%%%%%%%%%%%%%%%%%%%%%%%%%%%%%%%%%%%%%%%%%%%%%%%%%%%%%%%%%%%%%%%%%%%
%%%%%%%%%%%%%%%%%%%%%%%%%%%%%%%%%%%%%%%%%%%%%%%%%%%%%%%%%%%%%%%%%%%%%%%%%%%%%%%%
%%% anthropic-claude-sonnet-3.7
%%%%%%%%%%%%%%%%%%%%%%%%%%%%%%%%%%%%%%%%%%%%%%%%%%%%%%%%%%%%%%%%%%%%%%%%%%%%%%%%
%%%%%%%%%%%%%%%%%%%%%%%%%%%%%%%%%%%%%%%%%%%%%%%%%%%%%%%%%%%%%%%%%%%%%%%%%%%%%%%%
\subsubsection{Semantic and Structural Metrics - Claude 3.7 Sonnet}
%%% FIGURE %%%
\begin{figure}[H]
\centering
\includegraphics[scale=0.205]{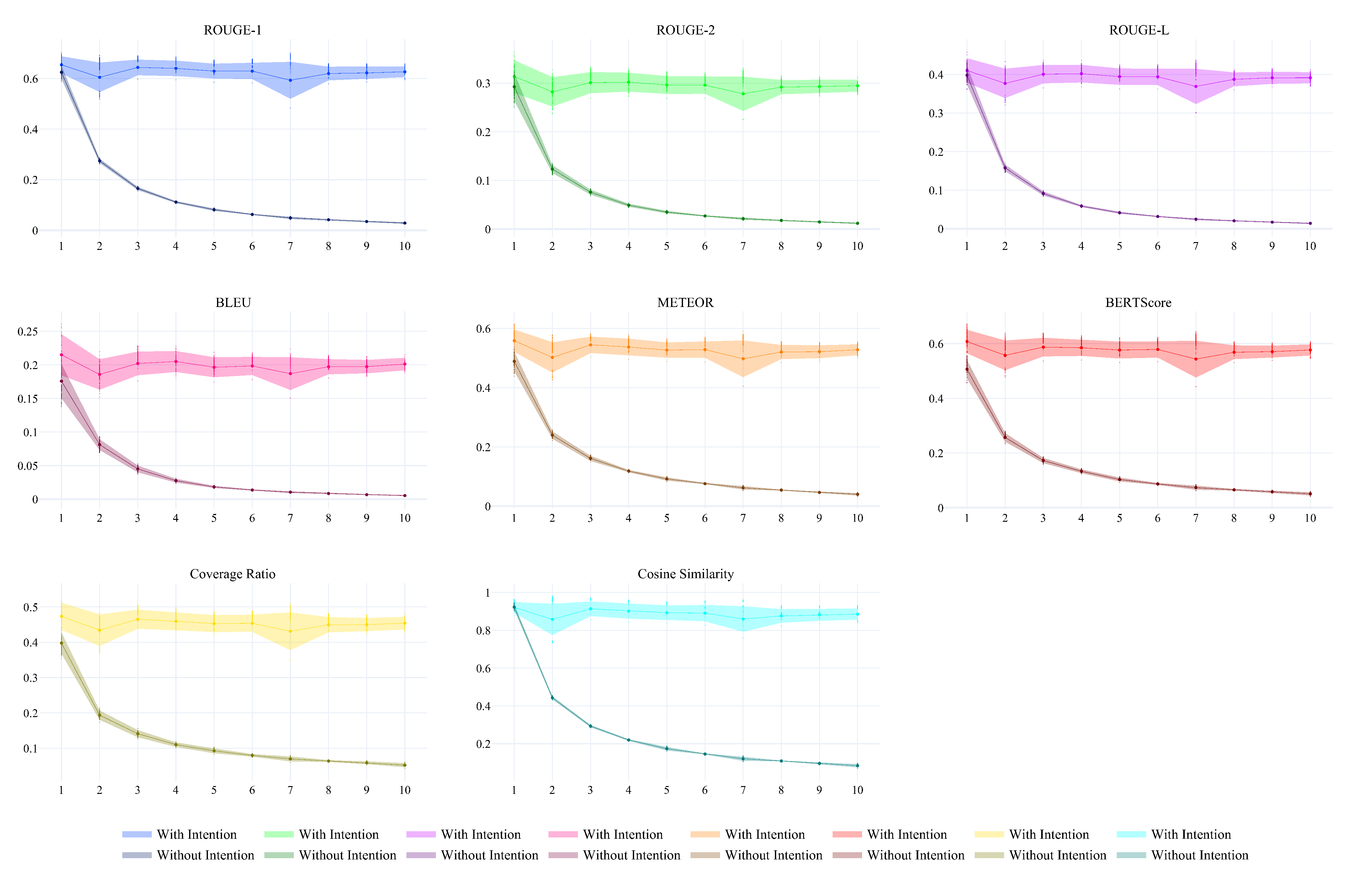}
\caption{\label{base_metrics_figure-anthropic-claude-sonnet-3.7} Semantic and Structural Metrics - Claude 3.7 Sonnet}
\end{figure}
%%% FIGURE %%%

%%% TABLE %%%
\begin{minipage}{\textwidth}
% \begin{table}[t]
\centering
\renewcommand{\arraystretch}{1.2}
\setlength{\tabcolsep}{8pt}
\resizebox{\textwidth}{!}{
\begin{tabular}{l | c c c c c c c c c c c}
\cmidrule(l){2-11}
\multicolumn{1}{l}{} & \multicolumn{10}{c}{Mixed Intention Level} \\
\hline
Metric & 1 & 2 & 3 & 4 & 5 & 6 & 7 & 8 & 9 & 10 \\
\hline

BLEU With Intention& \textbf{0.21498}& \textbf{0.18573}& \textbf{0.20215}& \textbf{0.20503}& \textbf{0.19644}& \textbf{0.19837}& \textbf{0.18693}& \textbf{0.19727}& \textbf{0.19738}& \textbf{0.20118} \\
BLEU Without Intention& 0.17584& 0.08125& 0.04503& 0.02757& 0.01824& 0.01369& 0.01042& 0.00847& 0.00669& 0.00542 \\

\hdashline

ROUGE-1 With Intention& \textbf{0.65420}& \textbf{0.60481}& \textbf{0.64384}& \textbf{0.64013}& \textbf{0.62939}& \textbf{0.62948}& \textbf{0.59299}& \textbf{0.61965}& \textbf{0.62213}& \textbf{0.62663} \\
ROUGE-1 Without Intention& 0.62471& 0.27454& 0.16520& 0.11134& 0.08129& 0.06258& 0.04893& 0.04173& 0.03450& 0.02852 \\

\hdashline

ROUGE-2 With Intention& \textbf{0.31346}& \textbf{0.28200}& \textbf{0.30095}& \textbf{0.30201}& \textbf{0.29610}& \textbf{0.29582}& \textbf{0.27801}& \textbf{0.29183}& \textbf{0.29292}& \textbf{0.29465} \\
ROUGE-2 Without Intention& 0.29249& 0.12327& 0.07584& 0.04905& 0.03489& 0.02698& 0.02134& 0.01779& 0.01467& 0.01195 \\

\hdashline

ROUGE-L With Intention& \textbf{0.41033}& \textbf{0.37675}& \textbf{0.40070}& \textbf{0.40194}& \textbf{0.39420}& \textbf{0.39398}& \textbf{0.36890}& \textbf{0.38727}& \textbf{0.39124}& \textbf{0.39137} \\
ROUGE-L Without Intention& 0.39788& 0.15782& 0.09137& 0.05865& 0.04136& 0.03174& 0.02455& 0.02050& 0.01704& 0.01389 \\

\hdashline

METEOR With Intention& \textbf{0.55870}& \textbf{0.50222}& \textbf{0.54468}& \textbf{0.53725}& \textbf{0.52701}& \textbf{0.52858}& \textbf{0.49767}& \textbf{0.52051}& \textbf{0.52149}& \textbf{0.52808} \\
METEOR Without Intention& 0.48897& 0.23986& 0.16130& 0.11836& 0.09185& 0.07602& 0.06191& 0.05431& 0.04645& 0.04002 \\

\hdashline

BERTScore With Intention& \textbf{0.60756}& \textbf{0.55724}& \textbf{0.58735}& \textbf{0.58501}& \textbf{0.57662}& \textbf{0.57882}& \textbf{0.54354}& \textbf{0.56853}& \textbf{0.57084}& \textbf{0.57681} \\
BERTScore Without Intention& 0.50637& 0.25708& 0.17254& 0.13320& 0.10314& 0.08681& 0.07323& 0.06540& 0.05823& 0.05061 \\

\hdashline

Coverage Ratio With Intention& \textbf{0.47349}& \textbf{0.43423}& \textbf{0.46529}& \textbf{0.45931}& \textbf{0.45294}& \textbf{0.45388}& \textbf{0.43119}& \textbf{0.44962}& \textbf{0.44998}& \textbf{0.45412} \\
Coverage Ratio Without Intention& 0.39765& 0.19349& 0.14103& 0.11015& 0.09253& 0.07967& 0.06964& 0.06362& 0.05877& 0.05199 \\

\hdashline

Cosine Similarity With Intention& 0.92057& \textbf{0.85755}& \textbf{0.91343}& \textbf{0.90114}& \textbf{0.89268}& \textbf{0.89053}& \textbf{0.85950}& \textbf{0.87574}& \textbf{0.88190}& \textbf{0.88530} \\
Cosine Similarity Without Intention& \textbf{0.92296}& 0.44303& 0.29298& 0.21965& 0.17345& 0.14548& 0.12025& 0.10880& 0.09582& 0.08373 \\

\hline
\end{tabular}}

\captionof{table}{Semantic and Structural Metrics - Claude 3.7 Sonnet}
\label{tab:base_metrics_table-anthropic-claude-sonnet-3.7}
\end{minipage}

%%% TABLE %%%

\subsubsection{LLM-as-a-Judge Scores (scaled to [0,1]) - Claude 3.7 Sonnet}
%%% FIGURE %%%
\begin{figure}[H]
\centering
\includegraphics[scale=0.205]{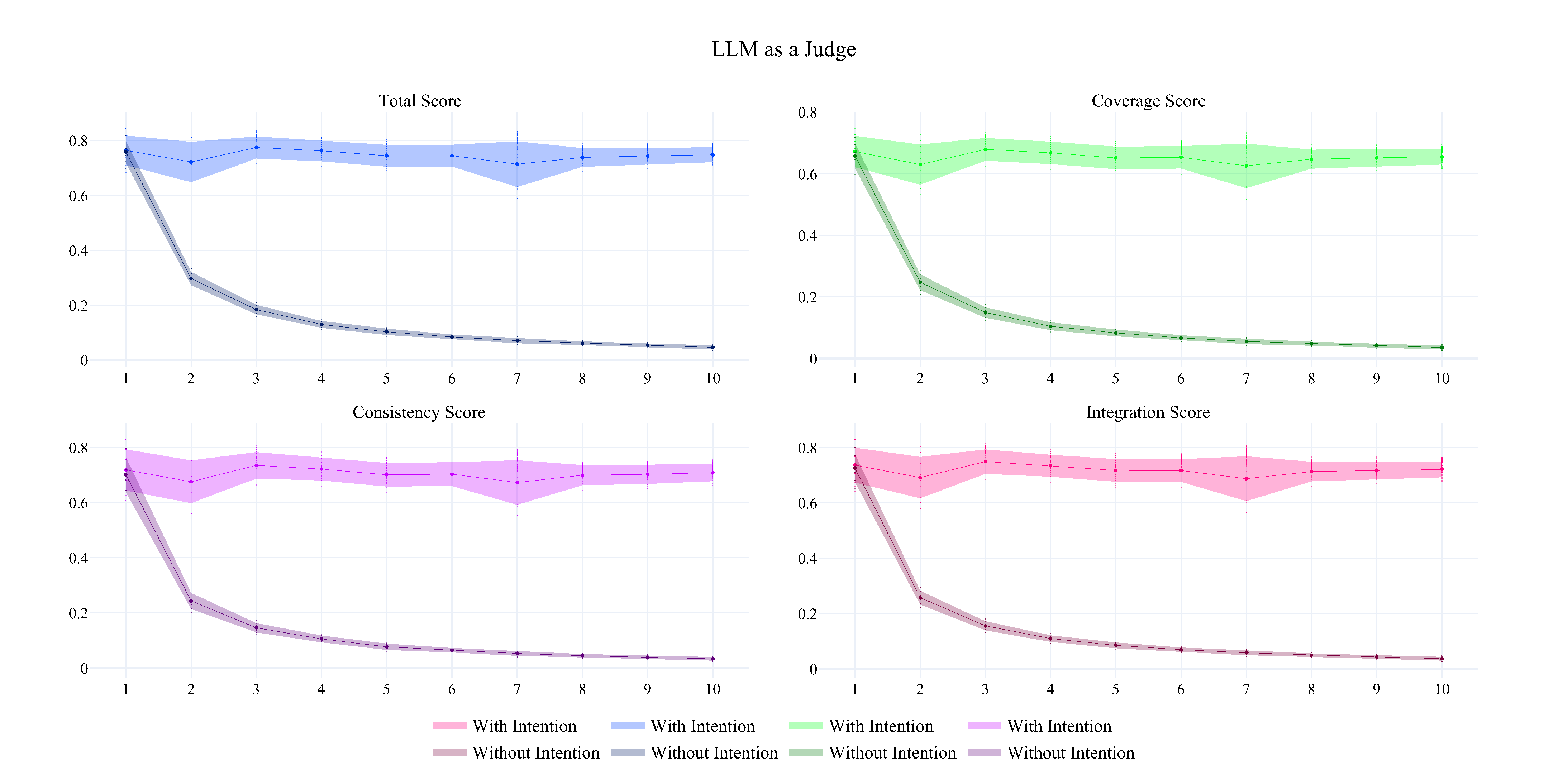}
\caption{\label{llm_as_a_judge_figure-anthropic-claude-sonnet-3.7} LLM-as-a-Judge Scores - Claude 3.7 Sonnet}
\end{figure}
%%% FIGURE %%%

%%% TABLE %%%
\begin{minipage}{\textwidth}
\centering
\renewcommand{\arraystretch}{1.2}
\setlength{\tabcolsep}{8pt}
\resizebox{\textwidth}{!}{
\begin{tabular}{l | c c c c c c c c c c c}
\cmidrule(l){2-11}
\multicolumn{1}{l}{} & \multicolumn{10}{c}{Mixed Intention Level} \\
\hline
Score & 1 & 2 & 3 & 4 & 5 & 6 & 7 & 8 & 9 & 10 \\
\hline

Coverage Score With Intention& \textbf{0.7645}& \textbf{0.7220}& \textbf{0.7750}& \textbf{0.7628}& \textbf{0.7448}& \textbf{0.7452}& \textbf{0.7141}& \textbf{0.7385}& \textbf{0.7441}& \textbf{0.7484} \\
Coverage Score Without Intention& 0.7590& 0.2970& 0.1837& 0.1295& 0.1026& 0.0838& 0.0703& 0.0609& 0.0534& 0.0458 \\

\hdashline

Consistency Score With Intention& \textbf{0.6720}& \textbf{0.6295}& \textbf{0.6790}& \textbf{0.6675}& \textbf{0.6512}& \textbf{0.6532}& \textbf{0.6253}& \textbf{0.6471}& \textbf{0.6516}& \textbf{0.6552} \\
Consistency Score Without Intention& 0.6580& 0.2475& 0.1493& 0.1048& 0.0830& 0.0668& 0.0554& 0.0480& 0.0420& 0.0356 \\

\hdashline

Integration Score With Intention& \textbf{0.7185}& \textbf{0.6755}& \textbf{0.7350}& \textbf{0.7215}& \textbf{0.7008}& \textbf{0.7030}& \textbf{0.6729}& \textbf{0.6994}& \textbf{0.7026}& \textbf{0.7083} \\
Integration Score Without Intention& 0.7010& 0.2440& 0.1463& 0.1060& 0.0774& 0.0652& 0.0534& 0.0450& 0.0393& 0.0340 \\

\hline

Total Score With Intention& \textbf{0.7365}& \textbf{0.6915}& \textbf{0.7493}& \textbf{0.7343}& \textbf{0.7174}& \textbf{0.7170}& \textbf{0.6876}& \textbf{0.7133}& \textbf{0.7176}& \textbf{0.7210} \\
Total Score Without Intention& 0.7260& 0.2575& 0.1557& 0.1095& 0.0850& 0.0695& 0.0581& 0.0495& 0.0433& 0.0368 \\

\hline
\end{tabular}}
\captionof{table}{LLM-as-a-Judge Scores - Claude 3.7 Sonnet}
\label{tab:llm_as_a_judge_table-anthropic-claude-sonnet-3.7}
\end{minipage}
%%% TABLE %%%

%%%%%%%%%%%%%%%%%%%%%%%%%%%%%%%%%%%%%%%%%%%%%%%%%%%%%%%%%%%%%%%%%%%%%%%%%%%%%%%%
%%%%%%%%%%%%%%%%%%%%%%%%%%%%%%%%%%%%%%%%%%%%%%%%%%%%%%%%%%%%%%%%%%%%%%%%%%%%%%%%
%%% openai-o1
%%%%%%%%%%%%%%%%%%%%%%%%%%%%%%%%%%%%%%%%%%%%%%%%%%%%%%%%%%%%%%%%%%%%%%%%%%%%%%%%
%%%%%%%%%%%%%%%%%%%%%%%%%%%%%%%%%%%%%%%%%%%%%%%%%%%%%%%%%%%%%%%%%%%%%%%%%%%%%%%%
\subsubsection{Semantic and Structural Metrics - OpenAI o1}
%%% FIGURE %%%
\begin{figure}[H]
\centering
\includegraphics[scale=0.205]{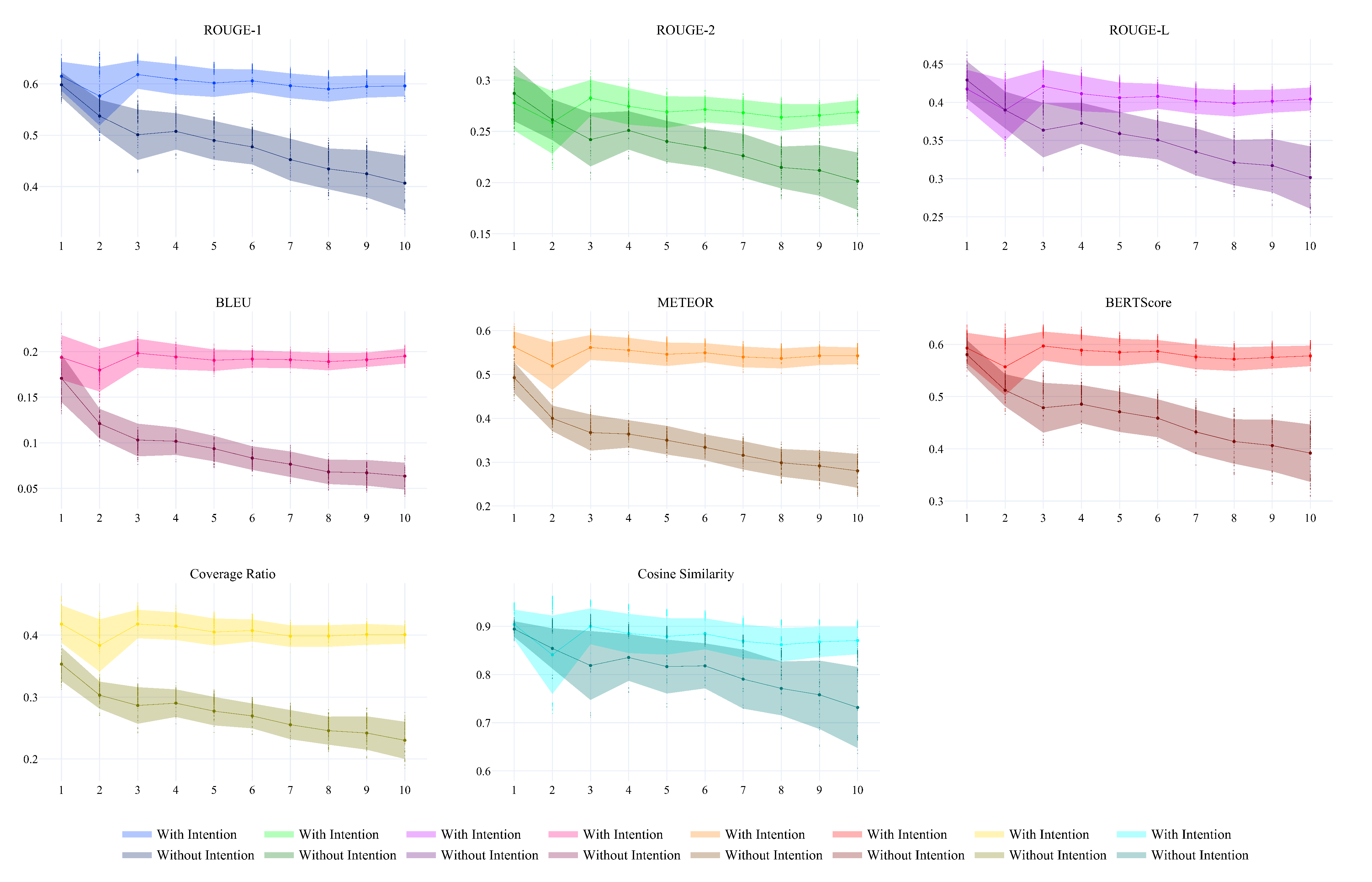}
\caption{\label{base_metrics_figure-openai-o1} Semantic and Structural Metrics - OpenAI o1}
\end{figure}
%%% FIGURE %%%

%%% TABLE %%%
\begin{minipage}{\textwidth}
% \begin{table}[t]
\centering
\renewcommand{\arraystretch}{1.2}
\setlength{\tabcolsep}{8pt}
\resizebox{\textwidth}{!}{ 
\begin{tabular}{l | c c c c c c c c c c c}
\cmidrule(l){2-11}
\multicolumn{1}{l}{} & \multicolumn{10}{c}{Mixed Intention Level} \\
\hline
Metric & 1 & 2 & 3 & 4 & 5 & 6 & 7 & 8 & 9 & 10 \\
\hline

BLEU With Intention& \textbf{0.1936}& \textbf{0.1797}& \textbf{0.1983}& \textbf{0.1940}& \textbf{0.1905}& \textbf{0.1918}& \textbf{0.1911}& \textbf{0.1890}& \textbf{0.1911}& \textbf{0.1951} \\
BLEU Without Intention& 0.1708& 0.1211& 0.1031& 0.1017& 0.0937& 0.0833& 0.0766& 0.0682& 0.0672& 0.0635 \\

\hdashline

ROUGE-1 With Intention& \textbf{0.6146}& \textbf{0.5760}& \textbf{0.6179}& \textbf{0.6084}& \textbf{0.6016}& \textbf{0.6057}& \textbf{0.5961}& \textbf{0.5897}& \textbf{0.5948}& \textbf{0.5959} \\
ROUGE-1 Without Intention& 0.5981& 0.5372& 0.5009& 0.5074& 0.4898& 0.4772& 0.4523& 0.4344& 0.4248& 0.4067 \\

\hdashline

ROUGE-2 With Intention& 0.2777& 0.2587& \textbf{0.2826}& \textbf{0.2745}& \textbf{0.2689}& \textbf{0.2714}& \textbf{0.2683}& \textbf{0.2637}& \textbf{0.2657}& \textbf{0.2689} \\
ROUGE-2 Without Intention& \textbf{0.2871}& \textbf{0.2614}& 0.2420& 0.2509& 0.2402& 0.2339& 0.2261& 0.2148& 0.2119& 0.2014 \\

\hdashline

ROUGE-L With Intention& 0.4173& 0.3896& \textbf{0.4210}& \textbf{0.4112}& \textbf{0.4059}& \textbf{0.4078}& \textbf{0.4016}& \textbf{0.3987}& \textbf{0.4014}& \textbf{0.4042} \\
ROUGE-L Without Intention& \textbf{0.4291}& \textbf{0.3899}& 0.3635& 0.3725& 0.3590& 0.3507& 0.3351& 0.3211& 0.3172& 0.3015 \\

\hdashline

METEOR With Intention& \textbf{0.5627}& \textbf{0.5193}& \textbf{0.5616}& \textbf{0.5554}& \textbf{0.5461}& \textbf{0.5497}& \textbf{0.5401}& \textbf{0.5367}& \textbf{0.5426}& \textbf{0.5426} \\
METEOR Without Intention& 0.4927& 0.4000& 0.3675& 0.3642& 0.3501& 0.3339& 0.3157& 0.2986& 0.2914& 0.2801 \\

\hdashline

BERTScore With Intention& \textbf{0.5929}& \textbf{0.5570}& \textbf{0.5970}& \textbf{0.5888}& \textbf{0.5848}& \textbf{0.5868}& \textbf{0.5761}& \textbf{0.5717}& \textbf{0.5751}& \textbf{0.5780} \\
BERTScore Without Intention& 0.5805& 0.5122& 0.4787& 0.4856& 0.4709& 0.4586& 0.4323& 0.4141& 0.4063& 0.3917 \\

\hdashline

Coverage Ratio With Intention& \textbf{0.4180}& \textbf{0.3834}& \textbf{0.4181}& \textbf{0.4146}& \textbf{0.4051}& \textbf{0.4076}& \textbf{0.3985}& \textbf{0.3986}& \textbf{0.4013}& \textbf{0.4009} \\
Coverage Ratio Without Intention& 0.3532& 0.3031& 0.2864& 0.2901& 0.2772& 0.2696& 0.2553& 0.2457& 0.2419& 0.2301 \\

\hdashline

Cosine Similarity With Intention& \textbf{0.9038}& 0.8410& \textbf{0.8999}& \textbf{0.8851}& \textbf{0.8790}& \textbf{0.8843}& \textbf{0.8691}& \textbf{0.8618}& \textbf{0.8676}& \textbf{0.8706} \\
Cosine Similarity Without Intention& 0.8943& \textbf{0.8541}& 0.8186& 0.8353& 0.8164& 0.8180& 0.7905& 0.7711& 0.7579& 0.7316 \\

\hline
\end{tabular}}

\captionof{table}{Semantic and Structural Metrics - OpenAI o1}
\label{tab:base_metrics_table-openai-o1}
% \end{table}
\end{minipage}

%%% TABLE %%%

\subsubsection{LLM-as-a-Judge Scores (scaled to [0,1]) - OpenAI o1}
%%% FIGURE %%%
\begin{figure}[H]
\centering
\includegraphics[scale=0.205]{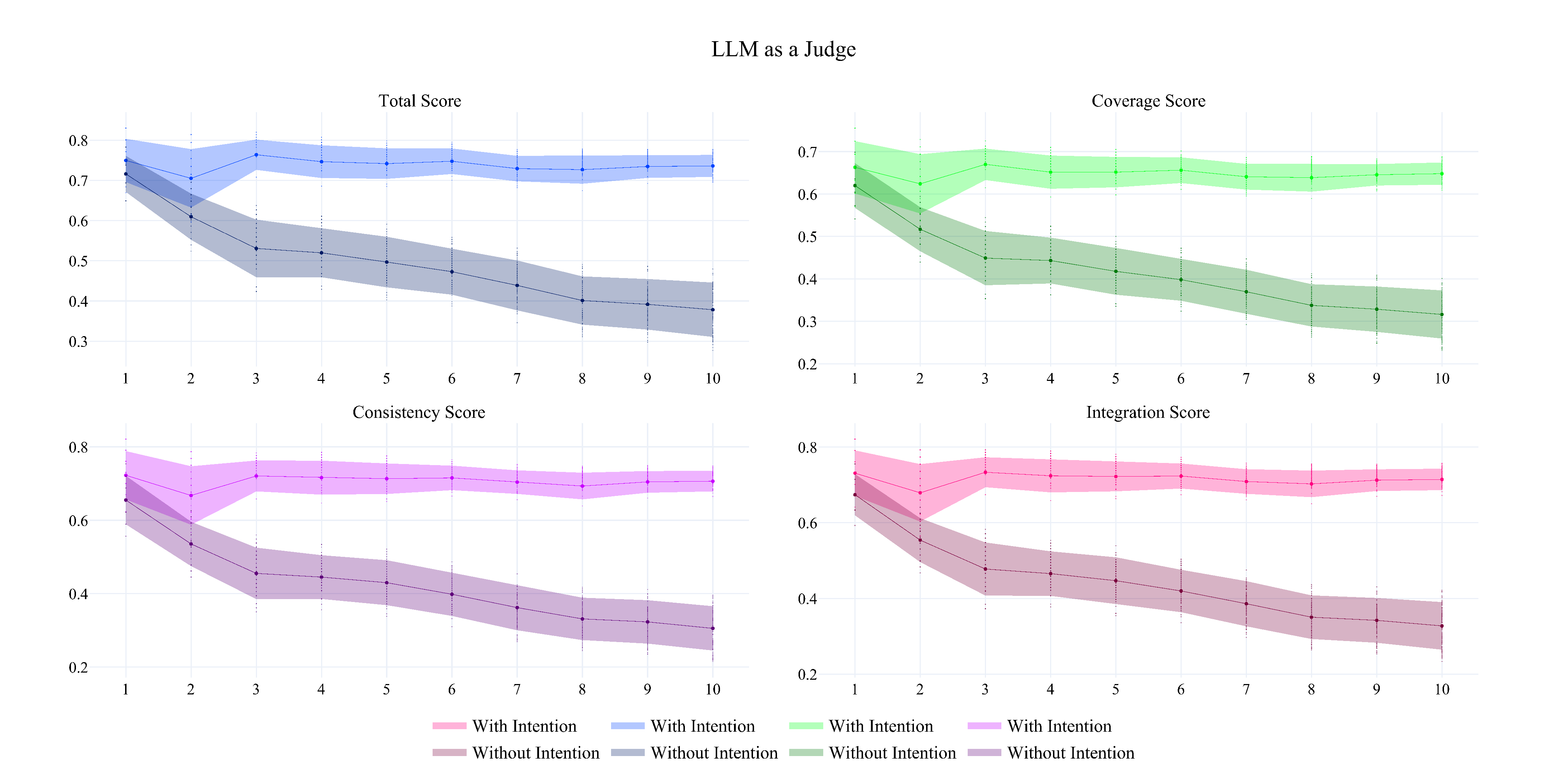}
\caption{\label{llm_as_a_judge_figure-openai-o1} LLM-as-a-Judge Scores - OpenAI o1}
\end{figure}
%%% FIGURE %%%

%%% TABLE %%%
\begin{minipage}{\textwidth}
\centering
\renewcommand{\arraystretch}{1.2}
\setlength{\tabcolsep}{8pt}
\resizebox{\textwidth}{!}{
\begin{tabular}{l | c c c c c c c c c c c}
\cmidrule(l){2-11}
\multicolumn{1}{l}{} & \multicolumn{10}{c}{Mixed Intention Level} \\
\hline
Score & 1 & 2 & 3 & 4 & 5 & 6 & 7 & 8 & 9 & 10 \\
\hline

Coverage Score With Intention& \textbf{0.7495}& \textbf{0.7050}& \textbf{0.7637}& \textbf{0.7465}& \textbf{0.7416}& \textbf{0.7475}& \textbf{0.7293}& \textbf{0.7266}& \textbf{0.7344}& \textbf{0.7359} \\
Coverage Score Without Intention& 0.7160& 0.6095& 0.5307& 0.5198& 0.4968& 0.4728& 0.4387& 0.4011& 0.3917& 0.3782 \\

\hdashline

Consistency Score With Intention& \textbf{0.6630}& \textbf{0.6240}& \textbf{0.6700}& \textbf{0.6515}& \textbf{0.6516}& \textbf{0.6562}& \textbf{0.6407}& \textbf{0.6386}& \textbf{0.6454}& \textbf{0.6480} \\
Consistency Score Without Intention& 0.6200& 0.5170& 0.4490& 0.4433& 0.4178& 0.3982& 0.3697& 0.3376& 0.3286& 0.3162 \\

\hdashline

Integration Score With Intention& \textbf{0.7225}& \textbf{0.6675}& \textbf{0.7207}& \textbf{0.7160}& \textbf{0.7132}& \textbf{0.7153}& \textbf{0.7039}& \textbf{0.6933}& \textbf{0.7046}& \textbf{0.7064} \\
Integration Score Without Intention& 0.6550& 0.5355& 0.4553& 0.4453& 0.4300& 0.3985& 0.3621& 0.3314& 0.3232& 0.3056 \\

\hline

Total Score With Intention& \textbf{0.7310}& \textbf{0.6790}& \textbf{0.7333}& \textbf{0.7240}& \textbf{0.7222}& \textbf{0.7233}& \textbf{0.7089}& \textbf{0.7026}& \textbf{0.7124}& \textbf{0.7144} \\
Total Score Without Intention& 0.6740& 0.5540& 0.4777& 0.4655& 0.4468& 0.4197& 0.3860& 0.3504& 0.3420& 0.3274 \\

\hline
\end{tabular}}
\captionof{table}{LLM-as-a-Judge Scores - OpenAI o1}
\label{tab:llm_as_a_judge_table-openai-o1}
\end{minipage}
%%% TABLE %%%

%%%%%%%%%%%%%%%%%%%%%%%%%%%%%%%%%%%%%%%%%%%%%%%%%%%%%%%%%%%%%%%%%%%%%%%%%%%%%%%%
%%%%%%%%%%%%%%%%%%%%%%%%%%%%%%%%%%%%%%%%%%%%%%%%%%%%%%%%%%%%%%%%%%%%%%%%%%%%%%%%
%%% openai-o3-mini
%%%%%%%%%%%%%%%%%%%%%%%%%%%%%%%%%%%%%%%%%%%%%%%%%%%%%%%%%%%%%%%%%%%%%%%%%%%%%%%%
%%%%%%%%%%%%%%%%%%%%%%%%%%%%%%%%%%%%%%%%%%%%%%%%%%%%%%%%%%%%%%%%%%%%%%%%%%%%%%%%
\subsubsection{Semantic and Structural Metrics - OpenAI o3-mini}
%%% FIGURE %%%
\begin{figure}[H]
\centering
\includegraphics[scale=0.205]{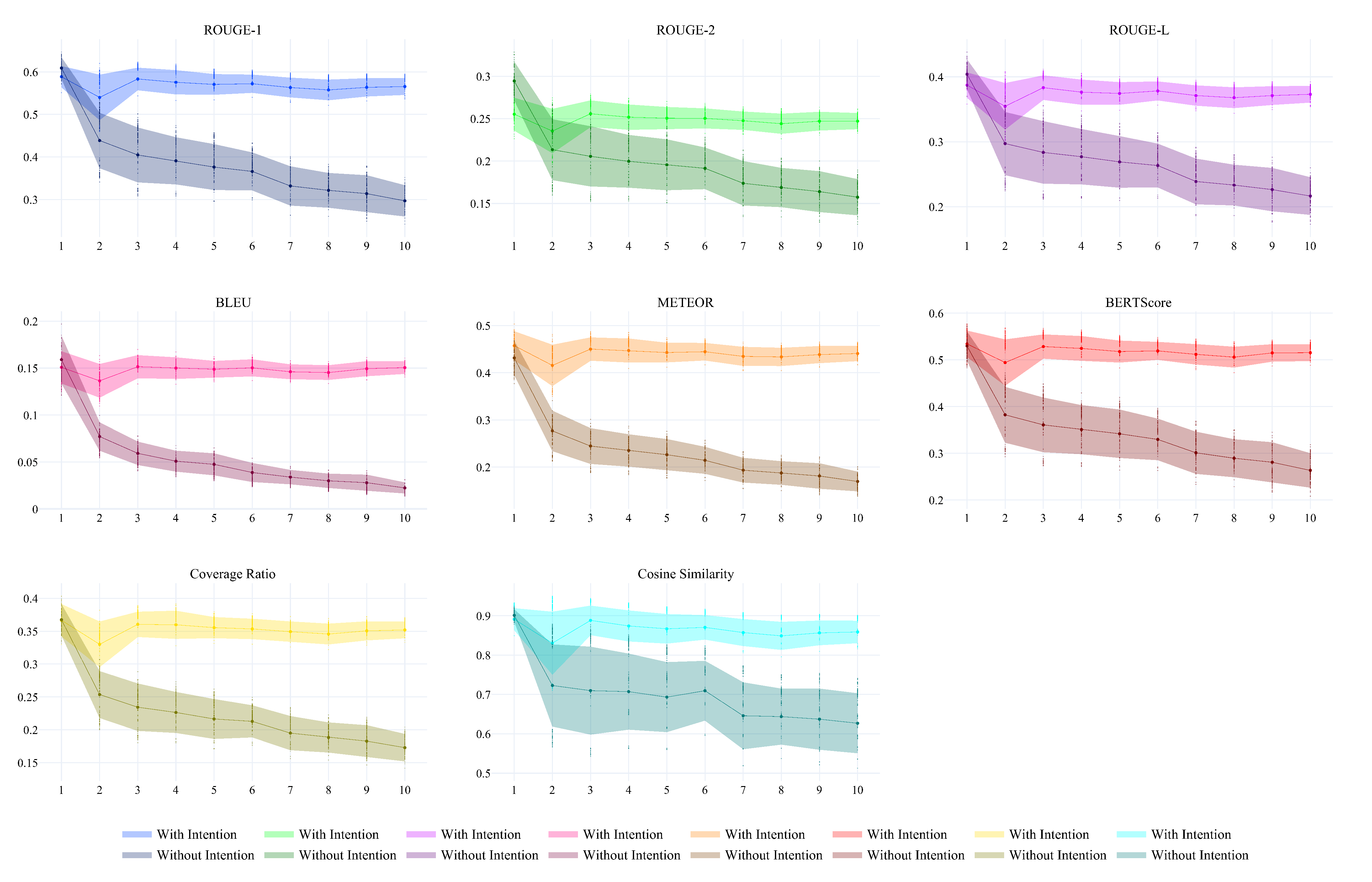}
\caption{\label{base_metrics_figure-openai-o3-mini} Semantic and Structural Metrics - OpenAI o3-mini}
\end{figure}
%%% FIGURE %%%

%%% TABLE %%%
\begin{minipage}{\textwidth}
% \begin{table}[t]
\centering
\renewcommand{\arraystretch}{1.2}
\setlength{\tabcolsep}{8pt}
\resizebox{\textwidth}{!}{
\begin{tabular}{l | c c c c c c c c c c c}
\cmidrule(l){2-11}
\multicolumn{1}{l}{} & \multicolumn{10}{c}{Mixed Intention Level} \\
\hline
Metric & 1 & 2 & 3 & 4 & 5 & 6 & 7 & 8 & 9 & 10 \\
\hline

BLEU With Intention& 0.1508& \textbf{0.1364}& \textbf{0.1515}& \textbf{0.1500}& \textbf{0.1488}& \textbf{0.1503}& \textbf{0.1461}& \textbf{0.1452}& \textbf{0.1494}& \textbf{0.1505} \\
BLEU Without Intention& \textbf{0.1589}& 0.0771& 0.0592& 0.0509& 0.0475& 0.0388& 0.0339& 0.0300& 0.0280& 0.0224 \\

\hdashline

ROUGE-1 With Intention& 0.5885& \textbf{0.5400}& \textbf{0.5833}& \textbf{0.5755}& \textbf{0.5705}& \textbf{0.5721}& \textbf{0.5631}& \textbf{0.5575}& \textbf{0.5636}& \textbf{0.5654} \\
ROUGE-1 Without Intention& \textbf{0.6091}& 0.4385& 0.4046& 0.3908& 0.3762& 0.3659& 0.3318& 0.3214& 0.3137& 0.2967 \\

\hdashline

ROUGE-2 With Intention& 0.2554& \textbf{0.2354}& \textbf{0.2559}& \textbf{0.2518}& \textbf{0.2506}& \textbf{0.2504}& \textbf{0.2476}& \textbf{0.2441}& \textbf{0.2471}& \textbf{0.2472} \\
ROUGE-2 Without Intention& \textbf{0.2945}& 0.2135& 0.2056& 0.1997& 0.1956& 0.1915& 0.1738& 0.1688& 0.1640& 0.1574 \\

\hdashline

ROUGE-L With Intention& 0.3872& \textbf{0.3546}& \textbf{0.3833}& \textbf{0.3764}& \textbf{0.3744}& \textbf{0.3783}& \textbf{0.3712}& \textbf{0.3678}& \textbf{0.3711}& \textbf{0.3731} \\
ROUGE-L Without Intention& \textbf{0.4040}& 0.2972& 0.2836& 0.2770& 0.2690& 0.2634& 0.2388& 0.2333& 0.2263& 0.2164 \\

\hdashline

METEOR With Intention& \textbf{0.4575}& \textbf{0.4155}& \textbf{0.4502}& \textbf{0.4472}& \textbf{0.4428}& \textbf{0.4446}& \textbf{0.4347}& \textbf{0.4336}& \textbf{0.4385}& \textbf{0.4408} \\
METEOR Without Intention& 0.4315& 0.2768& 0.2447& 0.2353& 0.2264& 0.2145& 0.1935& 0.1876& 0.1812& 0.1696 \\

\hdashline

BERTScore With Intention& \textbf{0.5345}& \textbf{0.4943}& \textbf{0.5286}& \textbf{0.5245}& \textbf{0.5178}& \textbf{0.5193}& \textbf{0.5119}& \textbf{0.5059}& \textbf{0.5149}& \textbf{0.5155} \\
BERTScore Without Intention& 0.5298& 0.3824& 0.3609& 0.3506& 0.3419& 0.3297& 0.3011& 0.2895& 0.2808& 0.2633 \\

\hdashline

Coverage Ratio With Intention& 0.3667& \textbf{0.3300}& \textbf{0.3604}& \textbf{0.3598}& \textbf{0.3554}& \textbf{0.3534}& \textbf{0.3493}& \textbf{0.3457}& \textbf{0.3505}& \textbf{0.3519} \\
Coverage Ratio Without Intention& \textbf{0.3675}& 0.2535& 0.2343& 0.2263& 0.2164& 0.2128& 0.1949& 0.1882& 0.1827& 0.1727 \\

\hdashline

Cosine Similarity With Intention& 0.8911& \textbf{0.8303}& \textbf{0.8882}& \textbf{0.8741}& \textbf{0.8668}& \textbf{0.8704}& \textbf{0.8573}& \textbf{0.8489}& \textbf{0.8564}& \textbf{0.8588} \\
Cosine Similarity Without Intention& \textbf{0.9011}& 0.7228& 0.7097& 0.7072& 0.6932& 0.7095& 0.6459& 0.6437& 0.6372& 0.6269 \\

\hline
\end{tabular}}

\captionof{table}{Semantic and Structural Metrics - OpenAI o3-mini}
\label{tab:base_metrics_table-openai-o3-mini}
% \end{table}
\end{minipage}
%%% TABLE %%%

\subsubsection{LLM-as-a-Judge Scores (scaled to [0,1]) - OpenAI o3-mini}
%%% FIGURE %%%
\begin{figure}[H]
\centering
\includegraphics[scale=0.205]{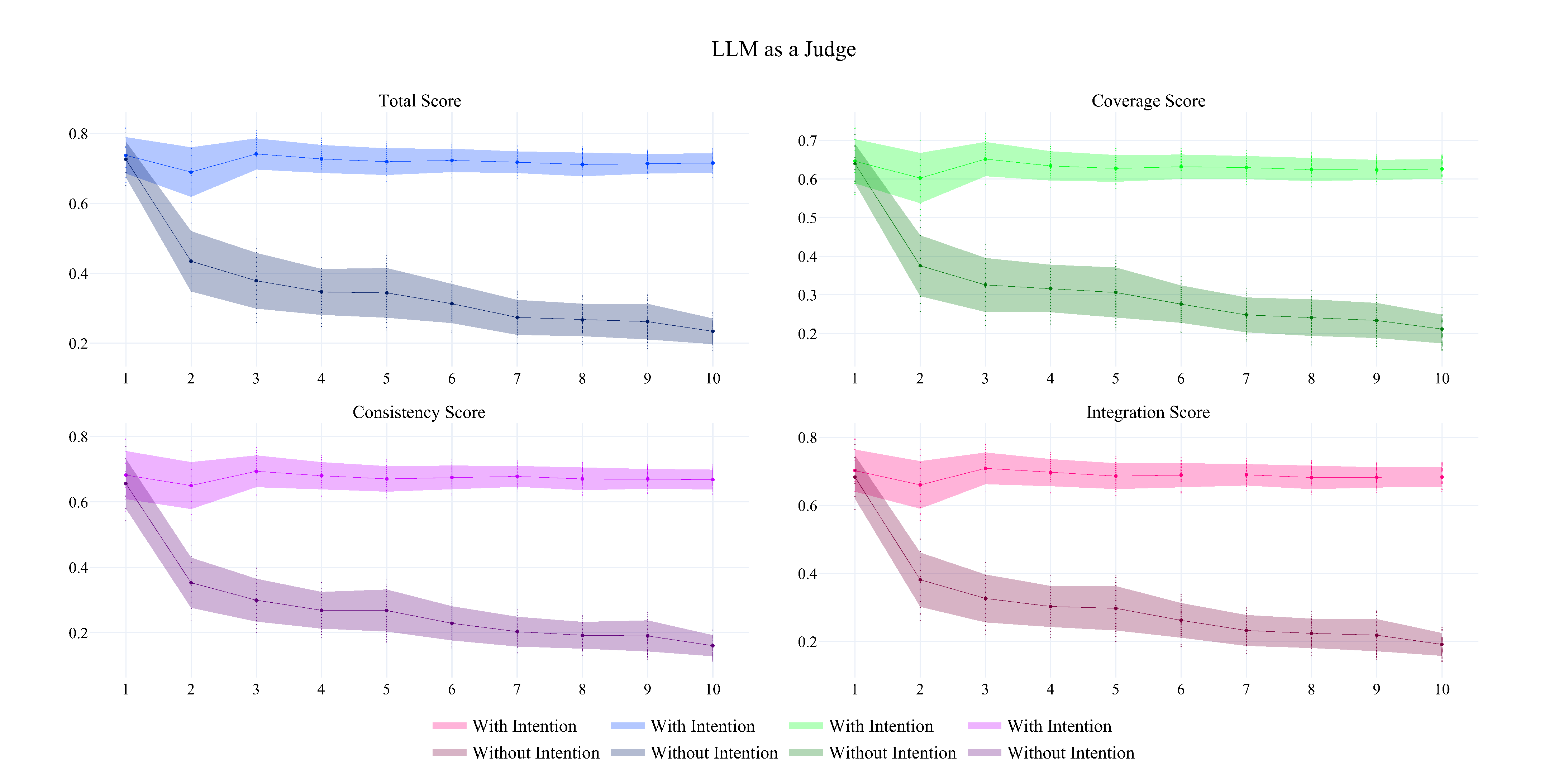}
\caption{\label{llm_as_a_judge_figure-openai-o3-mini} LLM-as-a-Judge Scores - OpenAI o3-mini}
\end{figure}
%%% FIGURE %%%

%%% TABLE %%%
\begin{minipage}{\textwidth}
\centering
\renewcommand{\arraystretch}{1.2}
\setlength{\tabcolsep}{8pt}
\resizebox{\textwidth}{!}{
\begin{tabular}{l | c c c c c c c c c c c}
\cmidrule(l){2-11}
\multicolumn{1}{l}{} & \multicolumn{10}{c}{Mixed Intention Level} \\
\hline
Score & 1 & 2 & 3 & 4 & 5 & 6 & 7 & 8 & 9 & 10 \\
\hline

Coverage Score With Intention& \textbf{0.7375}& \textbf{0.6895}& \textbf{0.7413}& \textbf{0.7270}& \textbf{0.7192}& \textbf{0.7228}& \textbf{0.7177}& \textbf{0.7111}& \textbf{0.7133}& \textbf{0.7154} \\
Coverage Score Without Intention& 0.7260& 0.4345& 0.3787& 0.3468& 0.3440& 0.3130& 0.2739& 0.2666& 0.2618& 0.2341 \\

\hdashline

Consistency Score With Intention& \textbf{0.6460}& \textbf{0.6025}& \textbf{0.6517}& \textbf{0.6340}& \textbf{0.6274}& \textbf{0.6318}& \textbf{0.6297}& \textbf{0.6243}& \textbf{0.6236}& \textbf{0.6263} \\
Consistency Score Without Intention& 0.6400& 0.3755& 0.3257& 0.3165& 0.3062& 0.2760& 0.2481& 0.2410& 0.2337& 0.2115 \\

\hdashline

Integration Score With Intention& \textbf{0.6820}& \textbf{0.6500}& \textbf{0.6937}& \textbf{0.6800}& \textbf{0.6704}& \textbf{0.6750}& \textbf{0.6780}& \textbf{0.6703}& \textbf{0.6704}& \textbf{0.6684} \\
Integration Score Without Intention& 0.6560& 0.3530& 0.2997& 0.2688& 0.2680& 0.2288& 0.2034& 0.1923& 0.1904& 0.1604 \\

\hline

Total Score With Intention& \textbf{0.7020}& \textbf{0.6600}& \textbf{0.7087}& \textbf{0.6960}& \textbf{0.6858}& \textbf{0.6888}& \textbf{0.6897}& \textbf{0.6818}& \textbf{0.6821}& \textbf{0.6831} \\
Total Score Without Intention& 0.6830& 0.3815& 0.3263& 0.3028& 0.2974& 0.2620& 0.2323& 0.2238& 0.2184& 0.1912 \\

\hline
\end{tabular}}
\captionof{table}{LLM-as-a-Judge Scores - OpenAI o3-mini}
\label{tab:llm_as_a_judge_table-openai-o3-mini}
\end{minipage}
%%% TABLE %%%

%%%%%%%%%%%%%%%%%%%%%%%%%%%%%%%%%%%%%%%%%%%%%%%%%%%%%%%%%%%%%%%%%%%%%%%%%%%%%%%%
%%%%%%%%%%%%%%%%%%%%%%%%%%%%%%%%%%%%%%%%%%%%%%%%%%%%%%%%%%%%%%%%%%%%%%%%%%%%%%%%
%%% openai-gpt-4.5
%%%%%%%%%%%%%%%%%%%%%%%%%%%%%%%%%%%%%%%%%%%%%%%%%%%%%%%%%%%%%%%%%%%%%%%%%%%%%%%%
%%%%%%%%%%%%%%%%%%%%%%%%%%%%%%%%%%%%%%%%%%%%%%%%%%%%%%%%%%%%%%%%%%%%%%%%%%%%%%%%
\subsubsection{Semantic and Structural Metrics - OpenAI GPT-4.5}
%%% FIGURE %%%
\begin{figure}[H]
\centering
\includegraphics[scale=0.2]{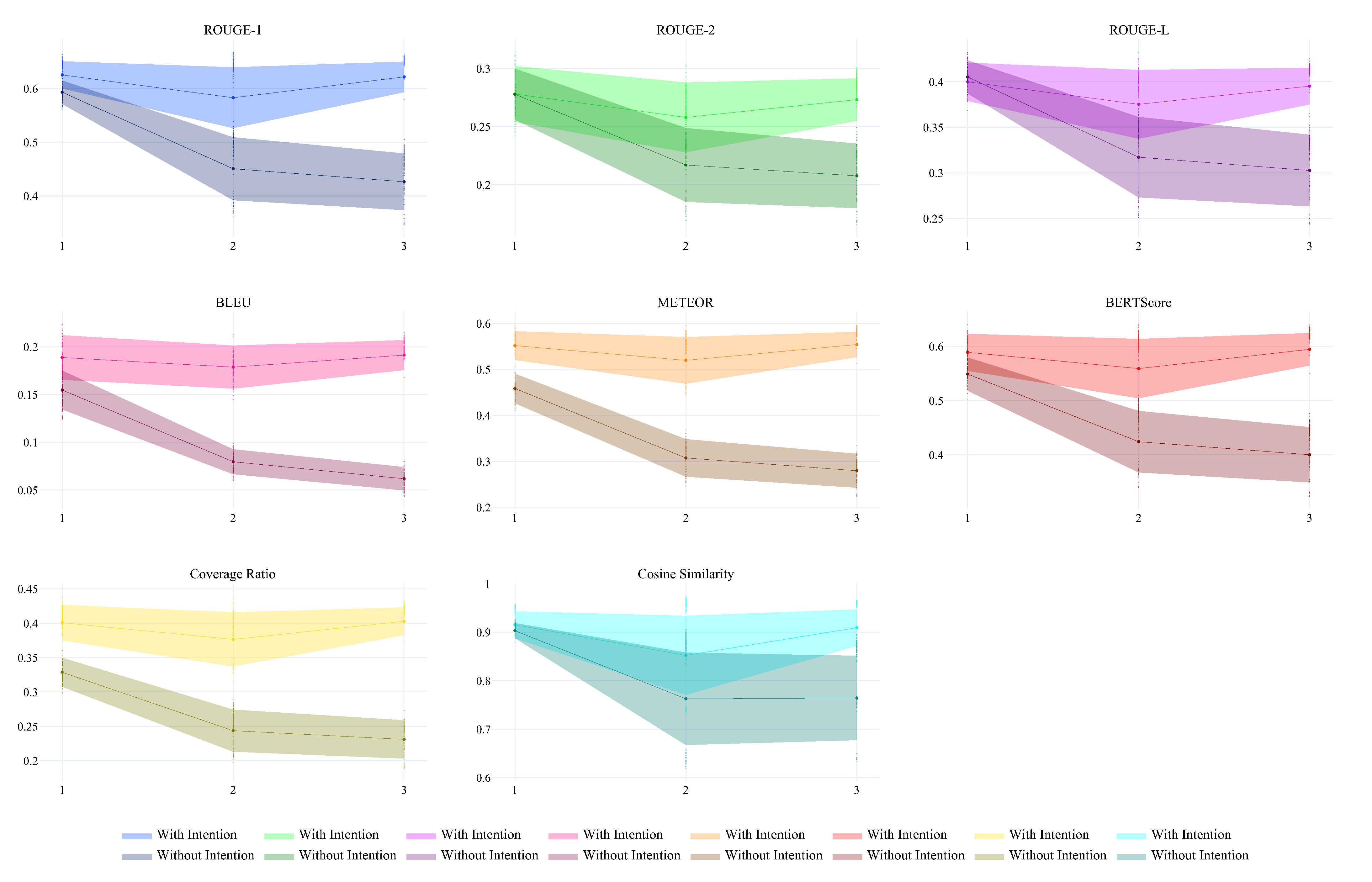}
\caption{\label{base_metrics_figure-openai-gpt-4.5} Semantic and Structural Metrics - OpenAI GPT-4.5}
\end{figure}
%%% FIGURE %%%

%%% TABLE %%%
\begin{minipage}{\textwidth}
    % \begin{table}[t]
    \centering
    \renewcommand{\arraystretch}{1.2}
    \setlength{\tabcolsep}{8pt}
    \resizebox{0.5\textwidth}{!}{
    \begin{tabular}{l | c c c }
    \cmidrule(l){2-4}
    \multicolumn{1}{l}{} & \multicolumn{3}{c}{Mixed Intention Level} \\
    \hline
    Metric & 1 & 2 & 3 \\
    \hline

BLEU With Intention & \textbf{0.1888} & \textbf{0.1787} & \textbf{0.1913} \\
BLEU Without Intention & 0.1547 & 0.0796 & 0.0618 \\
\hdashline
ROUGE-1 With Intention & \textbf{0.6252} & \textbf{0.5827} & \textbf{0.6214} \\
ROUGE-1 Without Intention & 0.5928 & 0.4506 & 0.4264 \\
\hdashline
ROUGE-2 With Intention & \textbf{0.2780} & \textbf{0.2578} & \textbf{0.2732} \\
ROUGE-2 Without Intention & 0.2778 & 0.2169 & 0.2075 \\
\hdashline
ROUGE-L With Intention & 0.4000 & \textbf{0.3752} & \textbf{0.3953} \\
ROUGE-L Without Intention & \textbf{0.4053} & 0.3171 & 0.3025 \\
\hdashline
METEOR With Intention & \textbf{0.5515} & \textbf{0.5195} & \textbf{0.5539} \\
METEOR Without Intention & 0.4583 & 0.3072 & 0.2794 \\
\hdashline
BERTScore With Intention & \textbf{0.5886} & \textbf{0.5588} & \textbf{0.5943} \\
BERTScore Without Intention & 0.5486 & 0.4242 & 0.4002 \\
\hdashline
Coverage Ratio With Intention & \textbf{0.4007} & \textbf{0.3764} & \textbf{0.4028} \\
Coverage Ratio Without Intention & 0.3286 & 0.2436 & 0.2309 \\
\hdashline
Cosine Similarity With Intention & \textbf{0.9152} & \textbf{0.8521} & \textbf{0.9093} \\
Cosine Similarity Without Intention & 0.9032 & 0.7624 & 0.7642 \\

\hline
\end{tabular}}

\captionof{table}{Semantic and Structural Metrics - OpenAI GPT-4.5}
\label{tab:base_metrics_table-openai-gpt-4.5}
\end{minipage}

%%% TABLE %%%

\subsubsection{LLM-as-a-Judge Scores (scaled to [0,1]) - OpenAI GPT-4.5}
%%% FIGURE %%%
\begin{figure}[H]
\centering
\includegraphics[scale=0.205]{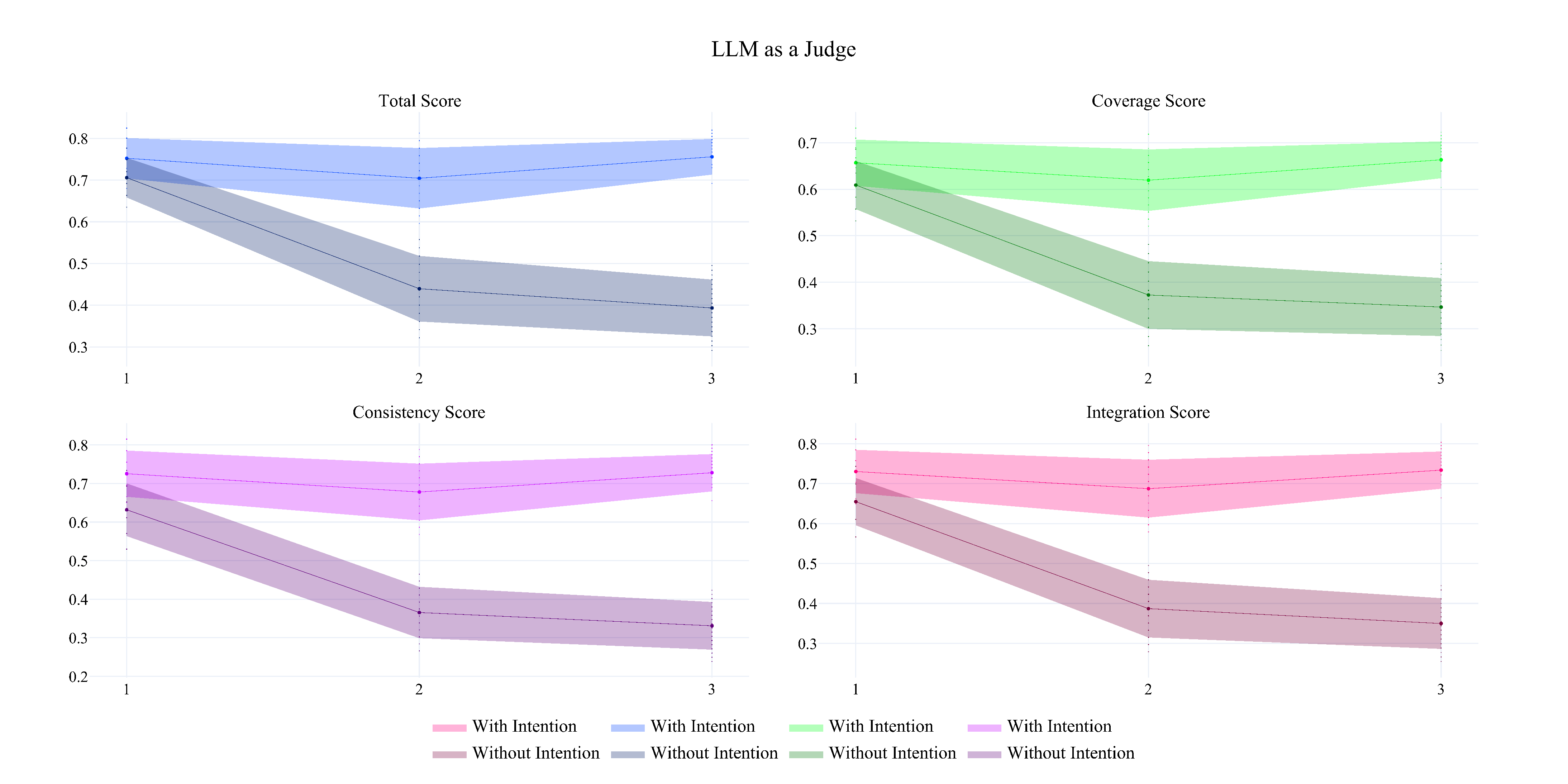}
\caption{\label{llm_as_a_judge_figure-openai-gpt-4.5} LLM-as-a-Judge Scores - OpenAI GPT-4.5}
\end{figure}
%%% FIGURE %%%

%%% TABLE %%%
\begin{minipage}{\textwidth}
\centering
\renewcommand{\arraystretch}{1.2}
\setlength{\tabcolsep}{8pt}
\resizebox{0.5\textwidth}{!}{
\begin{tabular}{l | c c c }
\cmidrule(l){2-4}
\multicolumn{1}{l}{} & \multicolumn{3}{c}{Mixed Intention Level} \\
\hline
Score & 1 & 2 & 3 \\
\hline

Coverage Score With Intention& \textbf{0.7525}& \textbf{0.7045}& \textbf{0.7560} \\
Coverage Score Without Intention& 0.7060& 0.4395& 0.3933 \\
\hdashline
Consistency Score With Intention& \textbf{0.6570}& \textbf{0.6195}& \textbf{0.6633} \\
Consistency Score Without Intention& 0.6090& 0.3725& 0.3467 \\
\hdashline
Integration Score With Intention& \textbf{0.7255}& \textbf{0.6780}& \textbf{0.7280} \\
Integration Score Without Intention& 0.6320& 0.3655& 0.3310 \\

\hline

Total Score With Intention& \textbf{0.7305}& \textbf{0.6875}& \textbf{0.7340} \\
Total Score Without Intention& 0.6550& 0.3870& 0.3497 \\

\hline
\end{tabular}}
\captionof{table}{LLM-as-a-Judge Scores - OpenAI GPT-4.5}
\label{tab:llm_as_a_judge_table-openai-gpt-4.5}
\end{minipage}

%%% TABLE %%%

%%%%%%%%%%%%%%%%%%%%%%%%%%%%%%%%%%%%%%%%%%%%%%%%%%%%%%%%%%%%%%%%%%%%%%%%%%%%%%%%
%%%%%%%%%%%%%%%%%%%%%%%%%%%%%%%%%%%%%%%%%%%%%%%%%%%%%%%%%%%%%%%%%%%%%%%%%%%%%%%%
%%% openai-gpt-4o
%%%%%%%%%%%%%%%%%%%%%%%%%%%%%%%%%%%%%%%%%%%%%%%%%%%%%%%%%%%%%%%%%%%%%%%%%%%%%%%%
%%%%%%%%%%%%%%%%%%%%%%%%%%%%%%%%%%%%%%%%%%%%%%%%%%%%%%%%%%%%%%%%%%%%%%%%%%%%%%%%
\subsubsection{Semantic and Structural Metrics - OpenAI GPT-4o}
%%% FIGURE %%%
\begin{figure}[H]
\centering
\includegraphics[scale=0.2]{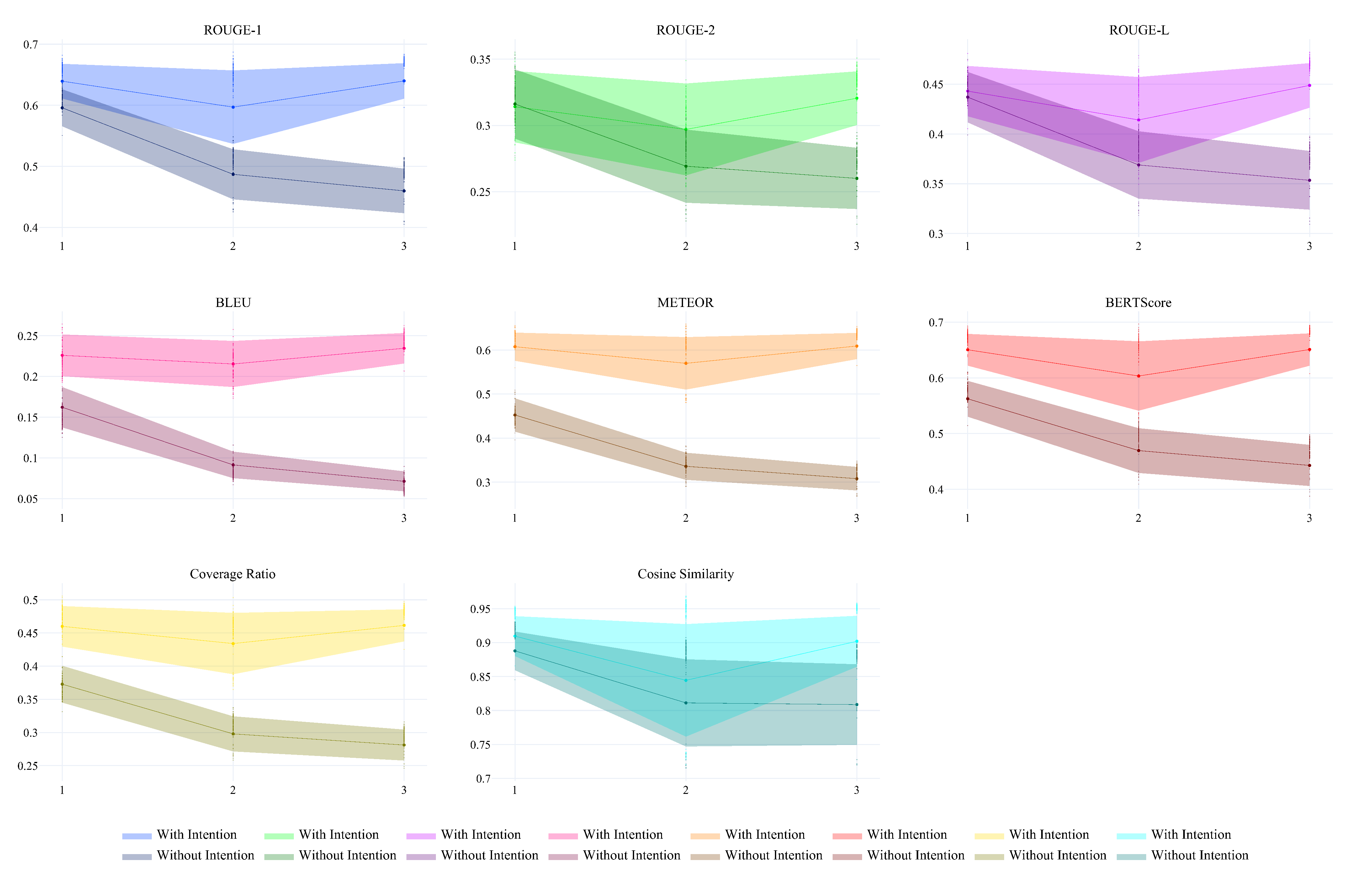}
\caption{\label{base_metrics_figure-openai-gpt-4o} Semantic and Structural Metrics - OpenAI GPT-4o}
\end{figure}
%%% FIGURE %%%

%%% TABLE %%%
\begin{minipage}{\textwidth}
\centering
\renewcommand{\arraystretch}{1.2}
\setlength{\tabcolsep}{8pt}
\resizebox{0.5\textwidth}{!}{
\begin{tabular}{l | c c c }
\cmidrule(l){2-4}
\multicolumn{1}{l}{} & \multicolumn{3}{c}{Mixed Intention Level} \\
\hline
Metric & 1 & 2 & 3 \\
\hline

BLEU With Intention & \textbf{0.2259} & \textbf{0.2153} & \textbf{0.2346} \\
BLEU Without Intention & 0.1622 & 0.0914 & 0.0712 \\
\hdashline
ROUGE-1 With Intention & \textbf{0.6395} & \textbf{0.5969} & \textbf{0.6399} \\
ROUGE-1 Without Intention & 0.5959 & 0.4869 & 0.4598 \\
\hdashline
ROUGE-2 With Intention & \textbf{0.3142} & \textbf{0.2970} & \textbf{0.3206} \\
ROUGE-2 Without Intention & 0.3162 & 0.2692 & 0.2600 \\
\hdashline
ROUGE-L With Intention & \textbf{0.4430} & \textbf{0.4141} & \textbf{0.4488} \\
ROUGE-L Without Intention & 0.4369 & 0.3689 & 0.3535 \\
\hdashline
METEOR With Intention & \textbf{0.6075} & \textbf{0.5698} & \textbf{0.6091} \\
METEOR Without Intention & 0.4525 & 0.3360 & 0.3079 \\
\hdashline
BERTScore With Intention & \textbf{0.6505} & \textbf{0.6034} & \textbf{0.6510} \\
BERTScore Without Intention & 0.5625 & 0.4695 & 0.4429 \\
\hdashline
Coverage Ratio With Intention & \textbf{0.4600} & \textbf{0.4340} & \textbf{0.4615} \\
Coverage Ratio Without Intention & 0.3728 & 0.2979 & 0.2811 \\
\hdashline
Cosine Similarity With Intention & \textbf{0.9096} & \textbf{0.8444} & \textbf{0.9020} \\
Cosine Similarity Without Intention & 0.8879 & 0.8112 & 0.8087 \\

\hline
\end{tabular}}

\captionof{table}{Semantic and Structural Metrics – OpenAI GPT-4o}
\label{tab:metrics_intention_vs_query}
\end{minipage}
%%% TABLE %%%

\subsubsection{LLM-as-a-Judge Scores (scaled to [0,1]) - OpenAI GPT-4o}
%%% FIGURE %%%
\begin{figure}[H]
\centering
\includegraphics[scale=0.205]{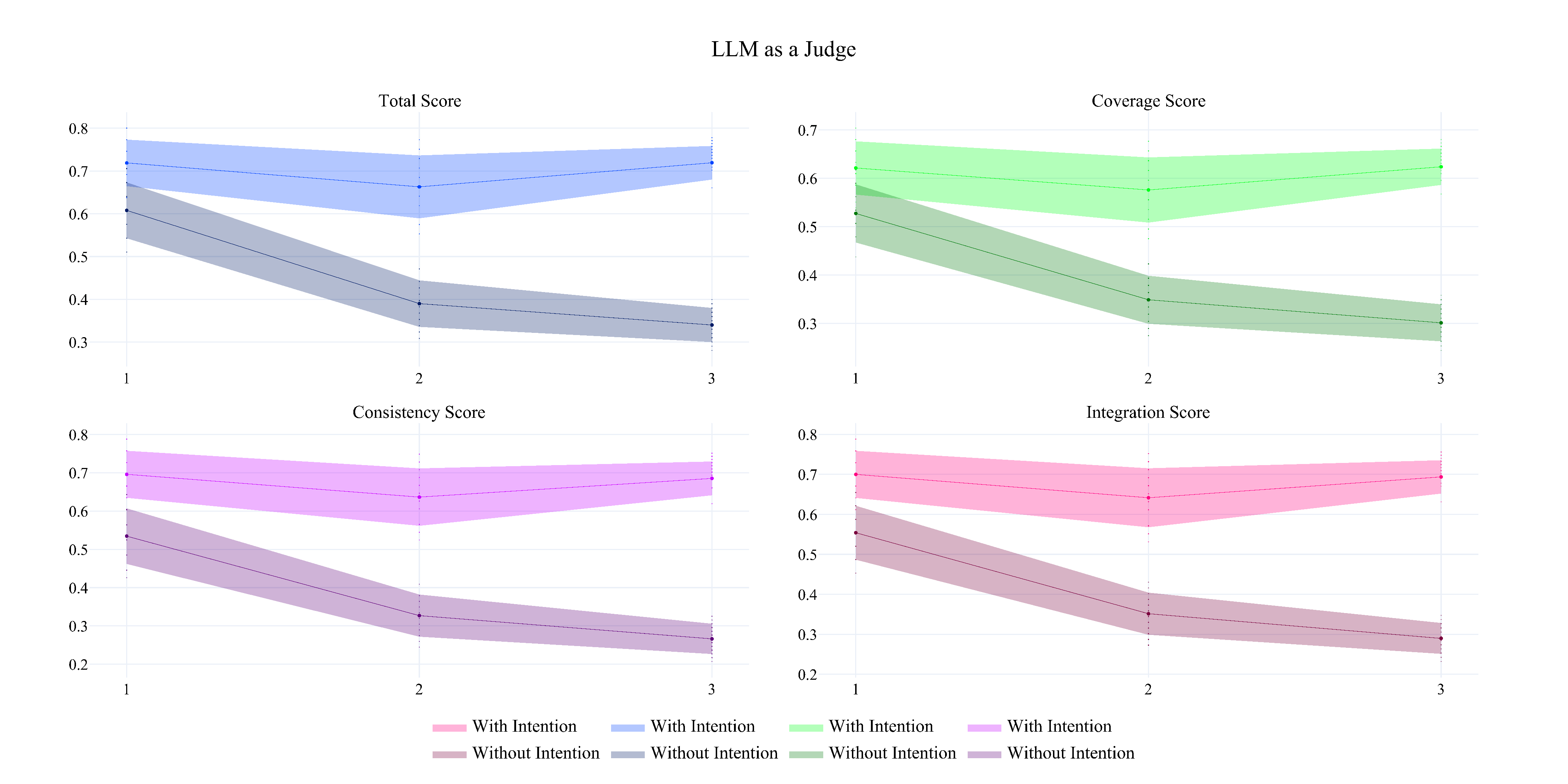}
\caption{\label{llm_as_a_judge_figure-openai-gpt-4o} LLM-as-a-Judge Scores - OpenAI GPT-4o}
\end{figure}
%%% FIGURE %%%

%%% TABLE %%%
\begin{minipage}{\textwidth}
\centering
\renewcommand{\arraystretch}{1.2}
\setlength{\tabcolsep}{8pt}
\resizebox{0.5\textwidth}{!}{
\begin{tabular}{l | c c c }
\cmidrule(l){2-4}
\multicolumn{1}{l}{} & \multicolumn{3}{c}{Mixed Intention Level} \\
\hline
Score & 1 & 2 & 3 \\
\hline

Coverage Score With Intention& \textbf{0.7190}& \textbf{0.6630}& \textbf{0.7193} \\
Coverage Score Without Intention& 0.6080& 0.3900& 0.3400 \\

\hdashline

Consistency Score With Intention& \textbf{0.6215}& \textbf{0.5760}& \textbf{0.6240} \\
Consistency Score Without Intention& 0.5275& 0.3490& 0.3013 \\

\hdashline

Integration Score With Intention& \textbf{0.6960}& \textbf{0.6365}& \textbf{0.6853} \\
Integration Score Without Intention& 0.5345& 0.3270& 0.2663 \\

\hline

Total Score With Intention& \textbf{0.7000}& \textbf{0.6415}& \textbf{0.6937} \\
Total Score Without Intention& 0.5540& 0.3515& 0.2897 \\

\hline
\end{tabular}}
\captionof{table}{LLM-as-a-Judge Scores - OpenAI GPT-4o}
\label{tab:llm_as_a_judge_table-openai-gpt-4o}
\end{minipage}
%%% TABLE %%%

%%%%%%%%%%%%%%%%%%%%%%%%%%%%%%%%%%%%%%%%%%%%%%%%%%%%%%%%%%%%%%%%%%%%%%%%%%%%%%%%
%%%%%%%%%%%%%%%%%%%%%%%%%%%%%%%%%%%%%%%%%%%%%%%%%%%%%%%%%%%%%%%%%%%%%%%%%%%%%%%%
%%% anthropic-claude-sonnet-3.5
%%%%%%%%%%%%%%%%%%%%%%%%%%%%%%%%%%%%%%%%%%%%%%%%%%%%%%%%%%%%%%%%%%%%%%%%%%%%%%%%
%%%%%%%%%%%%%%%%%%%%%%%%%%%%%%%%%%%%%%%%%%%%%%%%%%%%%%%%%%%%%%%%%%%%%%%%%%%%%%%%
\subsubsection{Semantic and Structural Metrics - Claude 3.5 Sonnet}
%%% FIGURE %%%
\begin{figure}[H]
\centering
\includegraphics[scale=0.205]{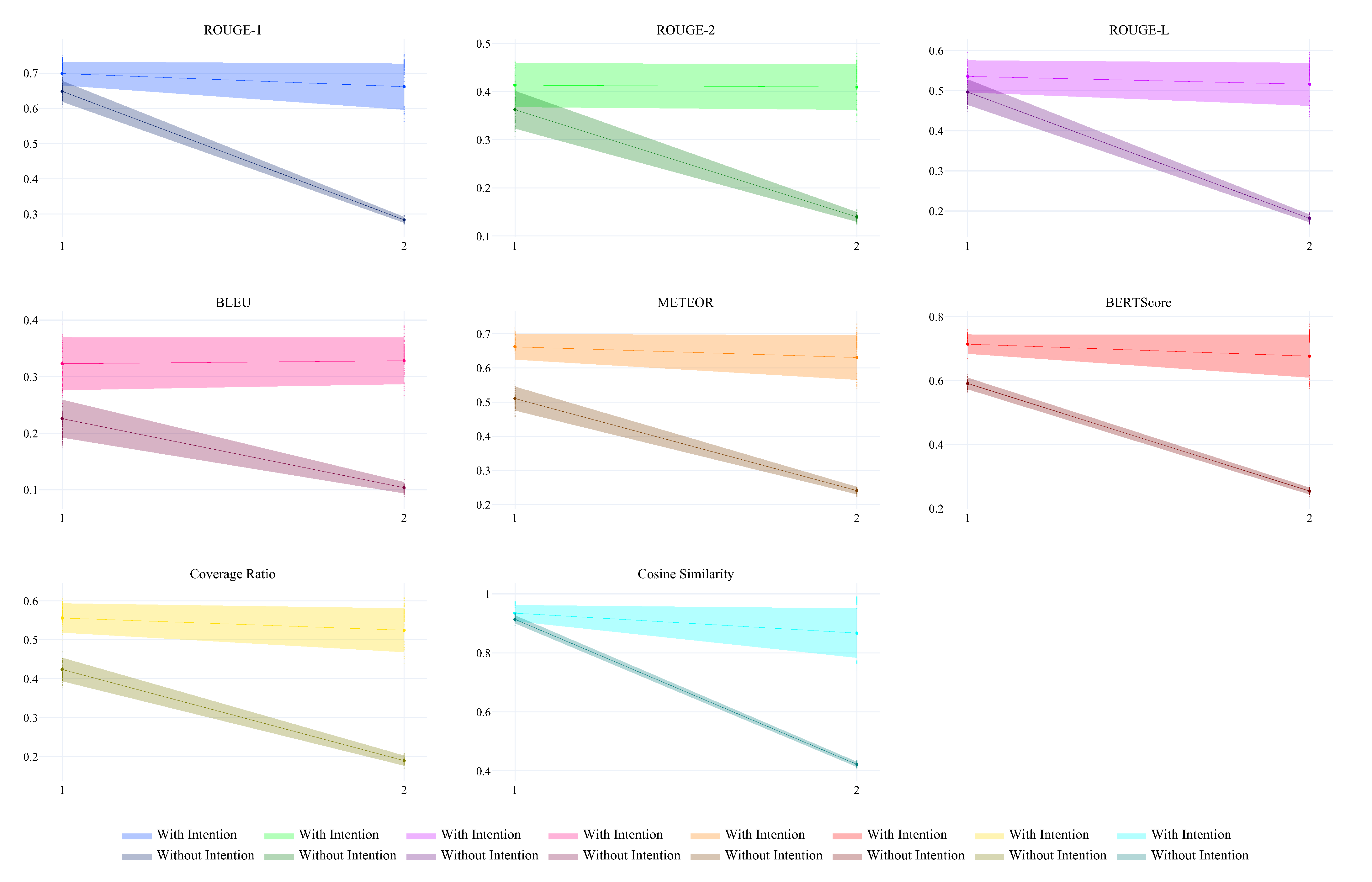}
\caption{\label{base_metrics_figure-anthropic-claude-sonnet-3.5} Semantic and Structural Metrics - Claude 3.5 Sonnet}
\end{figure}
%%% FIGURE %%%

%%% TABLE %%%
\begin{minipage}{\textwidth}
    % \begin{table}[t]
    \centering
    \renewcommand{\arraystretch}{1.2}
    \setlength{\tabcolsep}{8pt}
    \resizebox{0.4\textwidth}{!}{
    \begin{tabular}{l | c c }
    \cmidrule(l){2-3}
    \multicolumn{1}{l}{} & \multicolumn{2}{c}{\textbf{Mixed Intention Level}} \\
    \hline
    \textbf{Metric} & \textbf{1} & \textbf{2} \\
    \hline

BLEU With Intention            & \textbf{0.3229} & \textbf{0.3282} \\
BLEU Without Intention         & 0.2258 & 0.1036 \\

\hdashline

ROUGE-1 With Intention         & \textbf{0.6987} & \textbf{0.6613} \\
ROUGE-1 Without Intention      & 0.6482 & 0.2837 \\

\hdashline

ROUGE-2 With Intention         & \textbf{0.4133} & \textbf{0.4092} \\
ROUGE-2 Without Intention      & 0.3622 & 0.1395 \\

\hdashline

ROUGE-L With Intention         & \textbf{0.5355} & \textbf{0.5157} \\
ROUGE-L Without Intention      & 0.4963 & 0.1816 \\

\hdashline

METEOR With Intention           & \textbf{0.6617} & \textbf{0.6303} \\
METEOR Without Intention        & 0.5102 & 0.2405 \\

\hdashline

BERTScore With Intention        & \textbf{0.7139} & \textbf{0.6762} \\
BERTScore Without Intention     & 0.5908 & 0.2542 \\

\hdashline

Coverage Ratio With Intention   & \textbf{0.5561} & \textbf{0.5246} \\
Coverage Ratio Without Intention& 0.4238 & 0.1895 \\

\hdashline

Cosine Similarity With Intention & \textbf{0.9347} & \textbf{0.8671} \\
Cosine Similarity Without Intention& 0.9137 & 0.4225 \\

\hline
\end{tabular}}

\captionof{table}{Semantic and Structural Metrics - Claude 3.5 Sonnet}
\label{tab:base_metrics_table-anthropic-claude-sonnet-3.5}
% \end{table}
\end{minipage}
%%% TABLE %%%

\subsubsection{LLM-as-a-Judge Scores (scaled to [0,1]) - Claude 3.5 Sonnet}
%%% FIGURE %%%
\begin{figure}[H]
\centering
\includegraphics[scale=0.205]{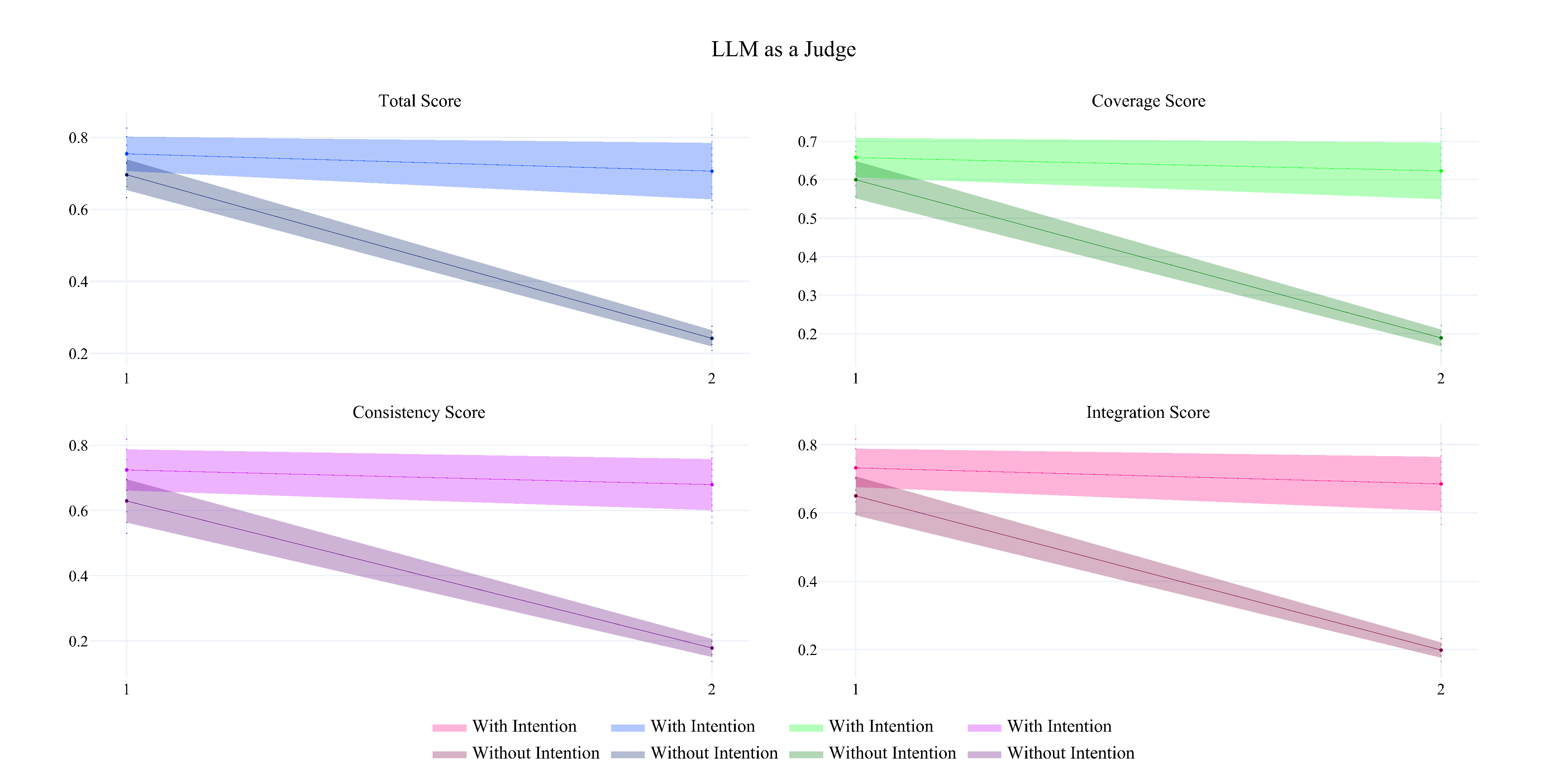}
\caption{\label{llm_as_a_judge_figure-anthropic-claude-sonnet-3.5} LLM-as-a-Judge Scores - Claude 3.5 Sonnet}
\end{figure}
%%% FIGURE %%%

%%% TABLE %%%
\begin{minipage}{\textwidth}
    \centering
    \renewcommand{\arraystretch}{1.2}
    \setlength{\tabcolsep}{8pt}
    \resizebox{0.4\textwidth}{!}{
    \begin{tabular}{l | c c }
    \cmidrule(l){2-3}
    \multicolumn{1}{l}{} & \multicolumn{2}{c}{Mixed Intention Level} \\
    \hline
    Score & \textbf{1} & \textbf{2} \\
    \hline

Coverage Score With Intention    & \textbf{0.7540} & \textbf{0.7060} \\
Coverage Score Without Intention & 0.6960 & 0.2415 \\

\hdashline

Consistency Score With Intention    & \textbf{0.6575} & \textbf{0.6230} \\
Consistency Score Without Intention & 0.6000 & 0.1890 \\

\hdashline

Integration Score With Intention    & \textbf{0.7240} & \textbf{0.6790} \\
Integration Score Without Intention & 0.6290 & 0.1785 \\

\hline

Total Score With Intention    & \textbf{0.7320} & \textbf{0.6850} \\
Total Score Without Intention & 0.6500 & 0.1985 \\

    \hline
    \end{tabular}}
    \captionof{table}{LLM-as-a-Judge Scores - Claude 3.5 Sonnet}
    \label{tab:llm_as_a_judge_table-anthropic-claude-sonnet-3.5}
\end{minipage}
%%% TABLE %%%

%%%%%%%%%%%%%%%%%%%%%%%%%%%%%%%%%%%%%%%%%%%%%%%%%%%%%%%%%%%%%%%%%%%%%%%%%%%%%%%%
%%%%%%%%%%%%%%%%%%%%%%%%%%%%%%%%%%%%%%%%%%%%%%%%%%%%%%%%%%%%%%%%%%%%%%%%%%%%%%%%
%%% deepseek-v3
%%%%%%%%%%%%%%%%%%%%%%%%%%%%%%%%%%%%%%%%%%%%%%%%%%%%%%%%%%%%%%%%%%%%%%%%%%%%%%%%
%%%%%%%%%%%%%%%%%%%%%%%%%%%%%%%%%%%%%%%%%%%%%%%%%%%%%%%%%%%%%%%%%%%%%%%%%%%%%%%%
\subsubsection{Semantic and Structural Metrics - DeepSeek-V3}
%%% FIGURE %%%
\begin{figure}[H]
\centering
\includegraphics[scale=0.205]{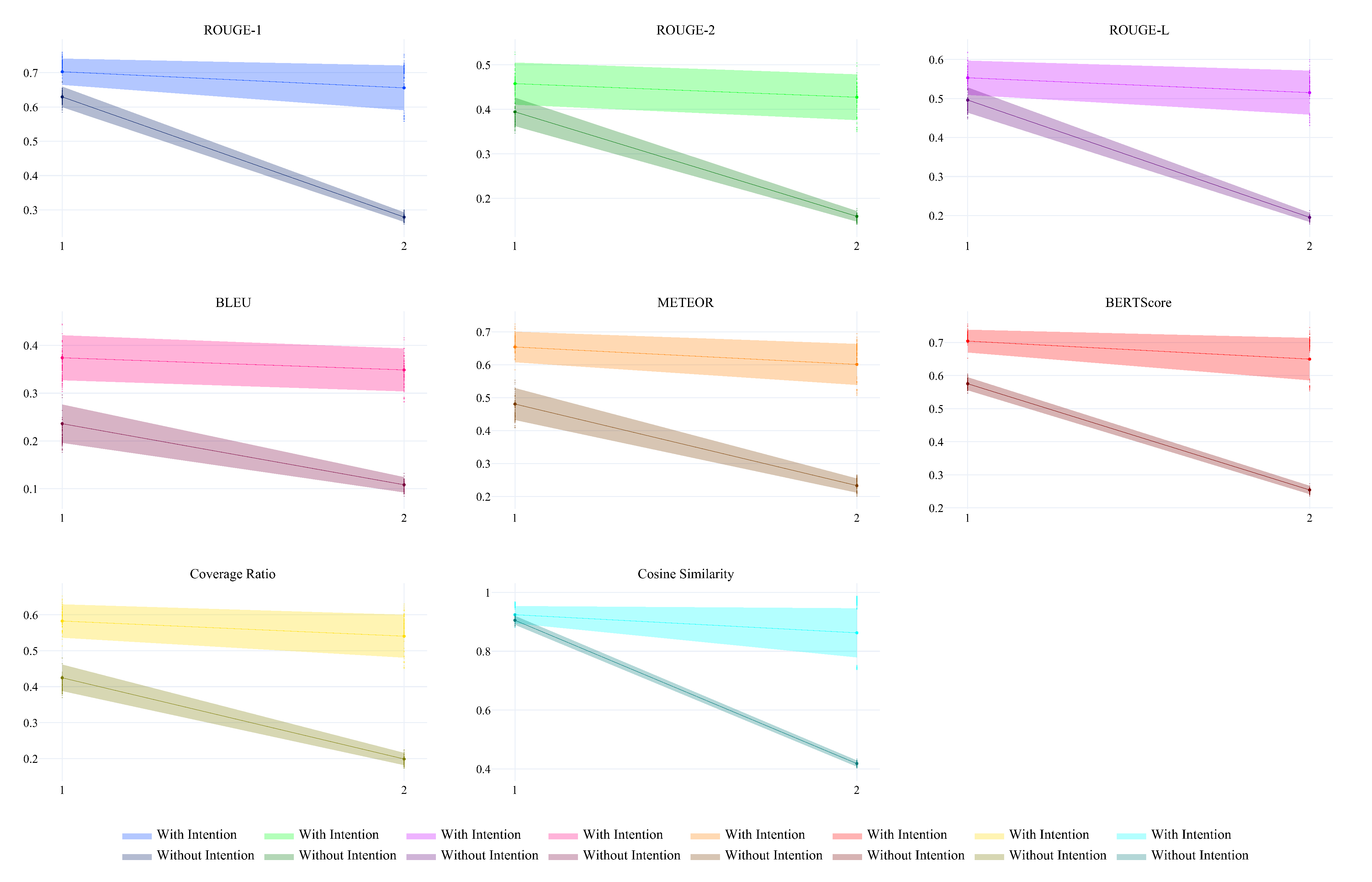}
\caption{\label{base_metrics_figure-deepseek-v3} Semantic and Structural Metrics - DeepSeek-V3}
\end{figure}
%%% FIGURE %%%

%%% TABLE %%%
\begin{minipage}{\textwidth}
% \begin{table}[t]
\centering
\renewcommand{\arraystretch}{1.2}
\setlength{\tabcolsep}{8pt}
\resizebox{0.4\textwidth}{!}{
\begin{tabular}{l | c c }
\cmidrule(l){2-3}
\multicolumn{1}{l}{} & \multicolumn{2}{c}{Mixed Intention Level} \\
\hline
Metric & 1 & 2 \\
\hline

BLEU With Intention& \textbf{0.3741}& \textbf{0.3486} \\
BLEU Without Intention& 0.2362& 0.1083 \\

\hdashline

ROUGE-1 With Intention& \textbf{0.7027}& \textbf{0.6557} \\
ROUGE-1 Without Intention& 0.6294& 0.2790 \\

\hdashline

ROUGE-2 With Intention& \textbf{0.4578}& \textbf{0.4272} \\
ROUGE-2 Without Intention& 0.3944& 0.1596 \\

\hdashline

ROUGE-L With Intention& \textbf{0.5532}& \textbf{0.5151} \\
ROUGE-L Without Intention& 0.4961& 0.1953 \\

\hdashline

METEOR With Intention& \textbf{0.6541}& \textbf{0.6010} \\
METEOR Without Intention& 0.4808& 0.2329 \\

\hdashline

BERTScore With Intention& \textbf{0.7041}& \textbf{0.6496} \\
BERTScore Without Intention& 0.5753& 0.2541 \\

\hdashline

Coverage Ratio With Intention& \textbf{0.5826}& \textbf{0.5405} \\
Coverage Ratio Without Intention& 0.4247& 0.1987 \\

\hdashline

Cosine Similarity With Intention& \textbf{0.9241}& \textbf{0.8625} \\
Cosine Similarity Without Intention& 0.9050& 0.4182 \\

\hline
\end{tabular}}

\captionof{table}{Semantic and Structural Metrics - DeepSeek-V3}
\label{tab:base_metrics_table-deepseek-v3}
% \end{table}
\end{minipage}

%%% TABLE %%%

\subsubsection{LLM-as-a-Judge Scores (scaled to [0,1]) - DeepSeek-V3}
%%% FIGURE %%%
\begin{figure}[H]
\centering
\includegraphics[scale=0.205]{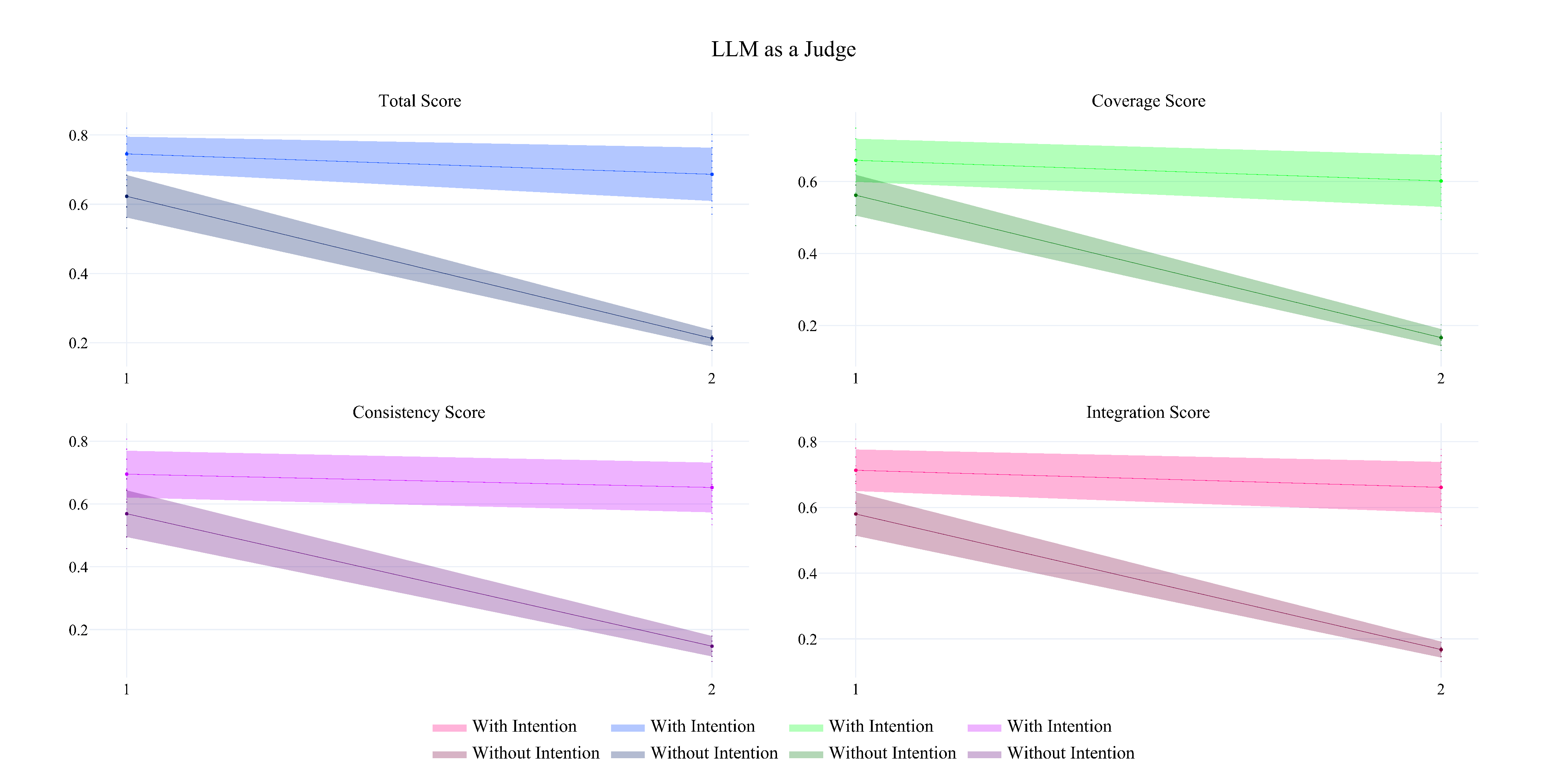}
\caption{\label{llm_as_a_judge_figure-deepseek-v3} LLM-as-a-Judge Scores - DeepSeek-V3}
\end{figure}
%%% FIGURE %%%

%%% TABLE %%%
\begin{minipage}{\textwidth}
\centering
\renewcommand{\arraystretch}{1.2}
\setlength{\tabcolsep}{8pt}
\resizebox{0.4\textwidth}{!}{
\begin{tabular}{l | c c c c c c c c c c c}
\cmidrule(l){2-3}
\multicolumn{1}{l}{} & \multicolumn{2}{c}{Mixed Intention Level} \\
\hline
Score & 1 & 2 \\
\hline

Coverage Score With Intention& \textbf{0.7455}& \textbf{0.6865} \\
Coverage Score Without Intention& 0.6230& 0.2125 \\

\hdashline

Consistency Score With Intention& \textbf{0.6590}& \textbf{0.6015} \\
Consistency Score Without Intention& 0.5620& 0.1660 \\

\hdashline

Integration Score With Intention& \textbf{0.6950}& \textbf{0.6525} \\
Integration Score Without Intention& 0.5690& 0.1470 \\

\hline

Total Score With Intention& \textbf{0.7130}& \textbf{0.6610} \\
Total Score Without Intention& 0.5800& 0.1675 \\

\hline
\end{tabular}}
\captionof{table}{LLM-as-a-Judge Scores - DeepSeek-V3}
\label{tab:llm_as_a_judge_table-deepseek-v3}
\end{minipage}
%%% TABLE %%%

%%%%%%%%%%%%%%%%%%%%%%%%%%%%%%%%%%%%%%%%%%%%%%%%%%%%%%%%%%%%%%%%%%%%%%%%%%%%%%%%
%%%%%%%%%%%%%%%%%%%%%%%%%%%%%%%%%%%%%%%%%%%%%%%%%%%%%%%%%%%%%%%%%%%%%%%%%%%%%%%%
%%% deepseek-r1
%%%%%%%%%%%%%%%%%%%%%%%%%%%%%%%%%%%%%%%%%%%%%%%%%%%%%%%%%%%%%%%%%%%%%%%%%%%%%%%%
%%%%%%%%%%%%%%%%%%%%%%%%%%%%%%%%%%%%%%%%%%%%%%%%%%%%%%%%%%%%%%%%%%%%%%%%%%%%%%%%
\subsubsection{Semantic and Structural Metrics - DeepSeek-R1}
%%% FIGURE %%%
\begin{figure}[H]
\centering
\includegraphics[scale=0.205]{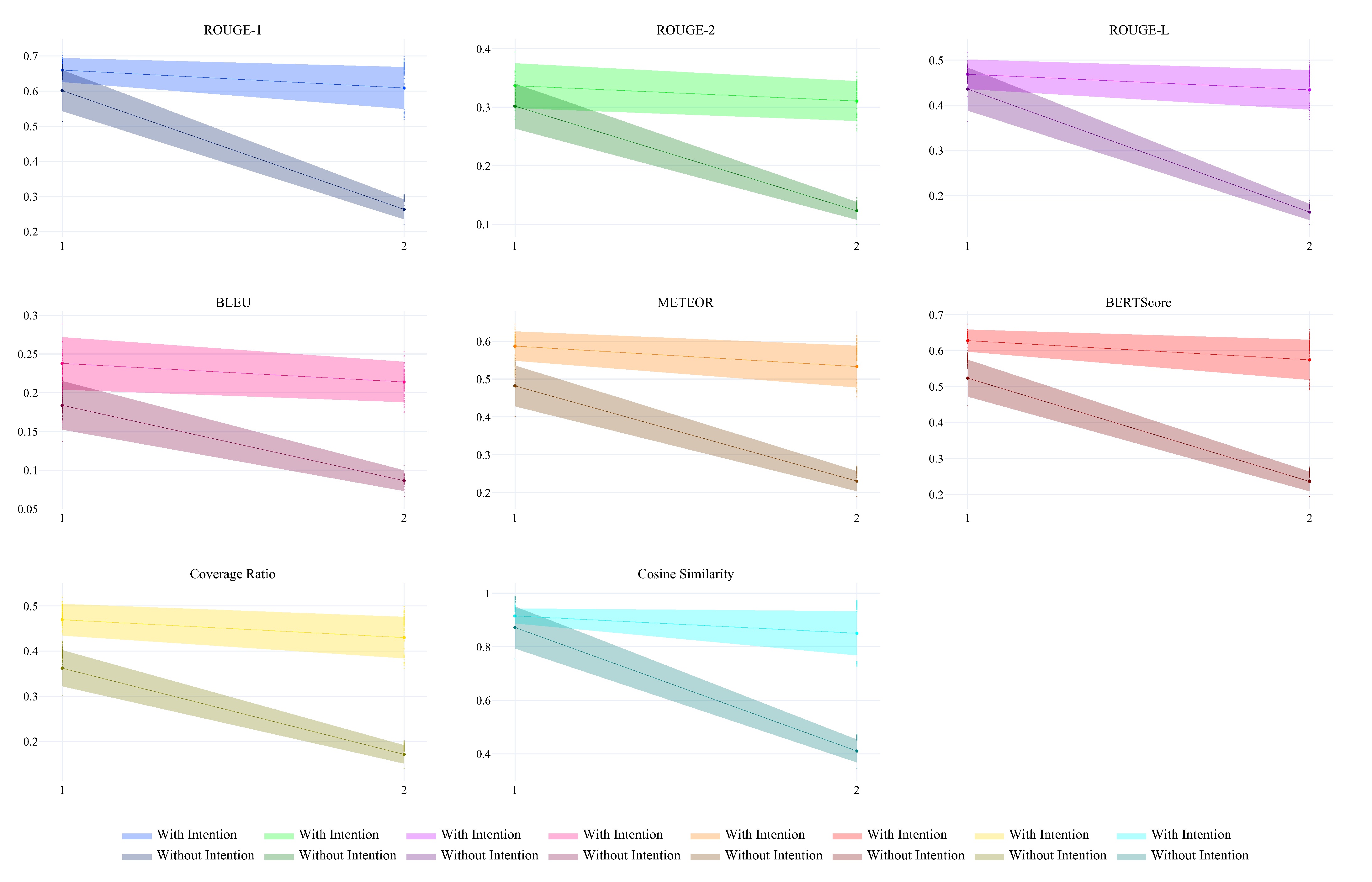}
\caption{\label{base_metrics_figure-deepseek-r1} Semantic and Structural Metrics - DeepSeek-R1}
\end{figure}
%%% FIGURE %%%

%%% TABLE %%%
\begin{minipage}{\textwidth}
% \begin{table}[t]
\centering
\renewcommand{\arraystretch}{1.2}
\setlength{\tabcolsep}{8pt}
\resizebox{0.4\textwidth}{!}{
\begin{tabular}{l | c c }
\cmidrule(l){2-3}
\multicolumn{1}{l}{} & \multicolumn{2}{c}{Mixed Intention Level} \\
\hline
Metric & 1 & 2 \\
\hline

BLEU With Intention& \textbf{0.2379}& \textbf{0.2139} \\
BLEU Without Intention& 0.1837& 0.0863 \\

\hdashline

ROUGE-1 With Intention& \textbf{0.6600}& \textbf{0.6088} \\
ROUGE-1 Without Intention& 0.6015& 0.2628 \\

\hdashline

ROUGE-2 With Intention& \textbf{0.3368}& \textbf{0.3106} \\
ROUGE-2 Without Intention& 0.3019& 0.1228 \\

\hdashline

ROUGE-L With Intention& \textbf{0.4686}& \textbf{0.4340} \\
ROUGE-L Without Intention& 0.4358& 0.1630 \\

\hdashline

METEOR With Intention& \textbf{0.5872}& \textbf{0.5331} \\
METEOR Without Intention& 0.4820& 0.2303 \\

\hdashline

BERTScore With Intention& \textbf{0.6274}& \textbf{0.5740} \\
BERTScore Without Intention& 0.5231& 0.2355 \\

\hdashline

Coverage Ratio With Intention& \textbf{0.4695}& \textbf{0.4299} \\
Coverage Ratio Without Intention& 0.3621& 0.1710 \\

\hdashline

Cosine Similarity With Intention& \textbf{0.9152}& \textbf{0.8504} \\
Cosine Similarity Without Intention& 0.8720& 0.4103 \\

\hline
\end{tabular}}

\captionof{table}{Semantic and Structural Metrics - DeepSeek-R1}
\label{tab:base_metrics_table-deepseek-r1}
% \end{table}
\end{minipage}
%%% TABLE %%%

\subsubsection{LLM-as-a-Judge Scores (scaled to [0,1]) - DeepSeek-R1}
%%% FIGURE %%%
\begin{figure}[H]
\centering
\includegraphics[scale=0.205]{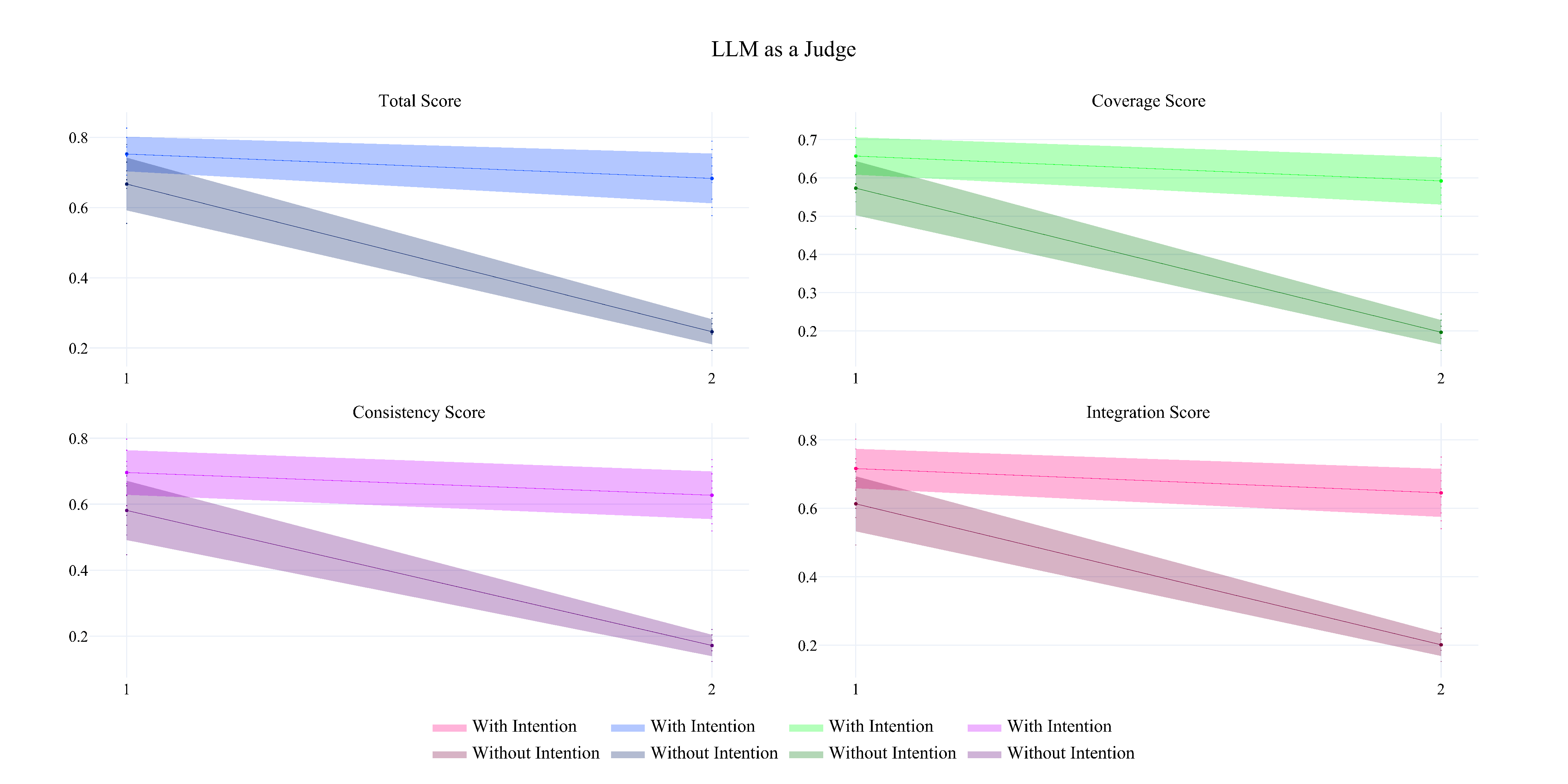}
\caption{\label{llm_as_a_judge_figure-deepseek-r1} LLM-as-a-Judge Scores - DeepSeek-R1}
\end{figure}
%%% FIGURE %%%

%%% TABLE %%%
\begin{minipage}{\textwidth}
\centering
\renewcommand{\arraystretch}{1.2}
\setlength{\tabcolsep}{8pt}
\resizebox{0.4\textwidth}{!}{
\begin{tabular}{l | c c }
\cmidrule(l){2-3}
\multicolumn{1}{l}{} & \multicolumn{2}{c}{Mixed Intention Level} \\
\hline
Score & 1 & 2 \\
\hline

Coverage Score With Intention& \textbf{0.7525}& \textbf{0.6830} \\
Coverage Score Without Intention& 0.6670& 0.2460 \\

\hdashline

Consistency Score With Intention& \textbf{0.6570}& \textbf{0.5920} \\
Consistency Score Without Intention& 0.5730& 0.1965 \\

\hdashline

Integration Score With Intention& \textbf{0.6960}& \textbf{0.6270} \\
Integration Score Without Intention& 0.5810& 0.1715 \\

\hline

Total Score With Intention& \textbf{0.7160}& \textbf{0.6450} \\
Total Score Without Intention& 0.6130& 0.2010 \\

\hline
\end{tabular}}
\captionof{table}{LLM-as-a-Judge Scores - DeepSeek-R1}
\label{tab:llm_as_a_judge_table-deepseek-r1}
\end{minipage}

%%% TABLE %%%

\end{document}